\documentclass[10pt,twocolumn,letterpaper]{article}

\usepackage{cvpr}              

\usepackage{xcolor}
\usepackage{colortbl}
\usepackage{fontawesome5}
\usepackage{multirow}
\usepackage{multicol}
\usepackage{fancyhdr}
\usepackage{marvosym}
\usepackage[most]{tcolorbox}
\usepackage{listings}
\usepackage{url}

\usepackage[final,commandnameprefix=always]{changes}
\setaddedmarkup{\textcolor{green!60!black}{#1}}
\setdeletedmarkup{}

\newcommand{\method}{\textsc{UniPixie}\xspace}
\newcommand{\dataset}{\textsc{PixieMultiVerse}\xspace}

\definecolor{cvprblue}{rgb}{0.21,0.49,0.74}
\definecolor{oursgreen}{RGB}{235, 250, 235}

\newcommand{\weburl}{\url{https://unipixie.github.io/}}

\lstdefinelanguage{json}{
    basicstyle=\ttfamily\footnotesize,
    numbers=left,
    numberstyle=\scriptsize\color{gray},
    stepnumber=1,
    numbersep=10pt,
    showstringspaces=false,
    breaklines=true,
    frame=none,
    backgroundcolor=\color{white},
    keywordstyle=\color{blue!70!black}\bfseries,
    stringstyle=\color{red!60!black},
    commentstyle=\color{green!50!black}\itshape,
    xleftmargin=15pt,
    xrightmargin=5pt
}

\newtcolorbox{promptbox}[1][]{
    colback=gray!5!white,
    colframe=gray!75!black,
    fonttitle=\bfseries,
    title=#1,
    enhanced,
    attach boxed title to top left={yshift=-2mm, xshift=2mm},
    boxrule=0.5pt,
    arc=2mm
}

\usepackage[pagebackref,breaklinks,colorlinks,allcolors=cvprblue]{hyperref}

\def\paperID{39186} 
\def\confName{CVPR}
\def\confYear{2026}

\title{UniPixie: Unified and Probabilistic 3D Physics Learning via Flow Matching

}

\author{Qilin Huang*$\dagger^{1,2}$, Quynh Anh Huynh*$^1$, Long Le*$^1$, Chen Wang$^1$, Chuhao Chen$^1$, \\
Ryan Lucas$^3$, Eric Eaton$^1$, Lingjie Liu$^1$ \\
$*$ Denotes equal contribution \quad $\dagger$ Work done during an internship at UPenn
\\
\vspace{-15pt}
\and
$^1$University of Pennsylvania \quad $^2$Southern University of Science and Technology \quad $^3$MIT \\
{\tt\small huangqilin2022@mail.sustech.edu.cn, ryanlu@mit.edu} \\
{\tt\small \{qanh308, vlongle, chenw30, chuhaoc, eeaton, lingjie.liu\}@seas.upenn.edu}
}

\begin{document}

\twocolumn[{%
\renewcommand\twocolumn[1][]{#1}%
\maketitle

\begin{center}
    \centering
    \vspace{-8pt}
    \includegraphics[width=0.98\textwidth]{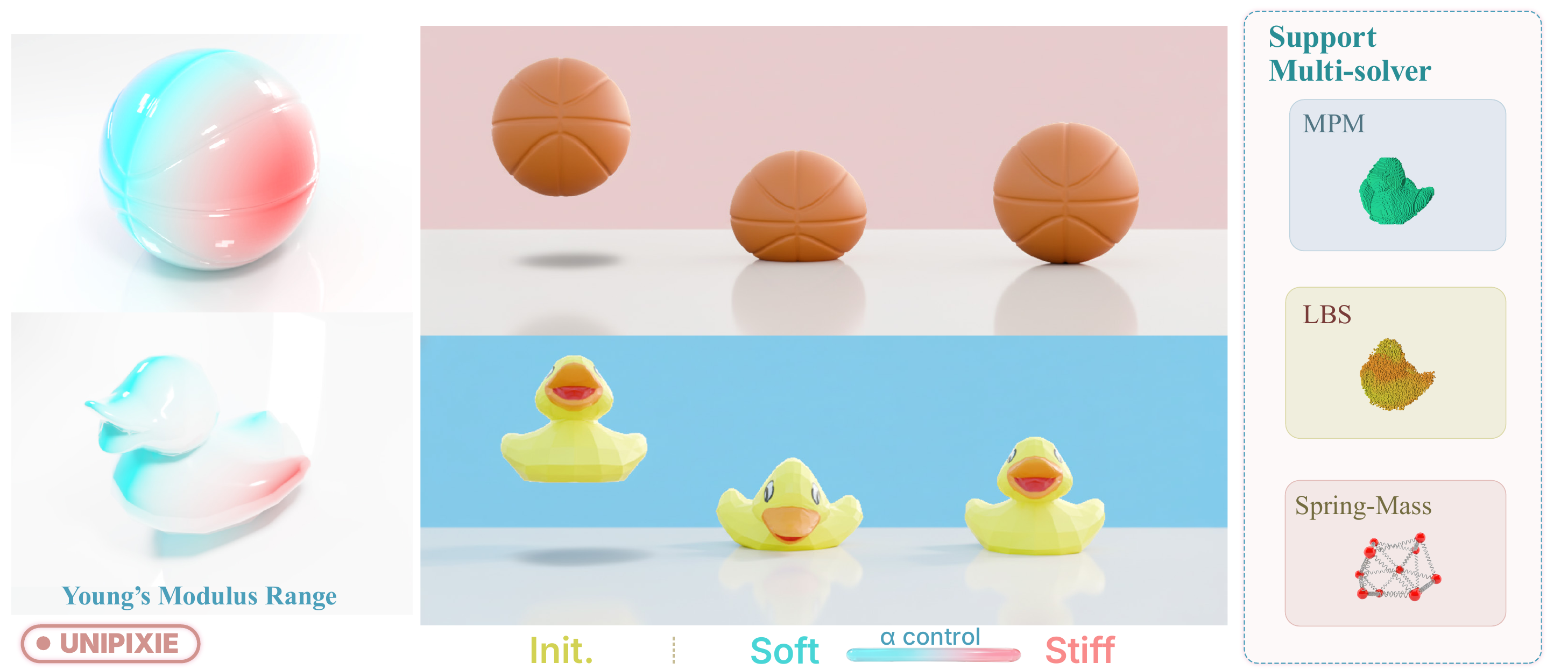}
    \vspace{-4pt}
    \captionof{figure}{
        \textbf{We introduce \method, a novel framework for controllable generation of a continuous range of physical properties from visual input.} Our model is trained on \dataset, a new dataset with annotated material property \textit{ranges}. The ground truth range for an object's Young's Modulus is visualized on the left, smoothly interpolating from its softest (blue) to stiffest (red) plausible value. By learning this continuous mapping, \method can generate a corresponding spectrum of dynamic behaviors from a single control parameter (center). A key innovation is our unified architecture's portability, producing consistent, simulation-ready parameters for diverse physics engines including Material Point Method (MPM), Linear Blend Skinning (LBS), and Spring-Mass systems (right).
    }
    \label{fig:hero}
\end{center}%
}]



\begin{abstract}
Existing feed-forward networks excel at predicting a single set of physical properties from visual appearance, but this point-estimate paradigm fundamentally fails to capture the real world's inherent physical ambiguity. We address this by reframing physics prediction as a task of learning a controllable, continuous distribution of material properties. We introduce \method, a framework trained to predict a continuous and parameterized path of physically plausible material properties from a single visual input. By learning a direct mapping along an object's softest-to-stiffest spectrum on our \dataset dataset, \method allows for controllable generation of diverse, physically-valid material fields via a single intuitive parameter. Crucially, \method introduces a novel unified architecture to produce simulation-ready parameters for diverse physics solvers, including continuum-based Material Point Method (MPM), reduced-order deformation based on Linear Blend Skinning (LBS), and anchor-based Spring-Mass systems, addressing a key portability issue in prior work. Experiments show our approach not only generates a rich variety of plausible dynamics but also \chreplaced{reduces Young's Modulus prediction error by over 50\% against the strongest deterministic baseline}{outperforms deterministic baselines by 2x}, bridging the gap between static point-estimates and the continuous nature of physical reality.
\weburl
\end{abstract}

\section{Introduction}
\vspace{-3pt}

Recent breakthroughs in 3D scene reconstruction, such as 3D Gaussian Splatting~\cite{kerbl20233d}, have enabled the creation of photorealistic digital replicas from images. Although these models can accurately capture the static appearance of our world, they remain oblivious to the underlying physics that govern {\em how} objects move, deform, and interact. A key frontier is therefore to infer an object’s material properties (e.g., Young’s modulus, density) directly from visual data. This “Physics-from-Pixels” task is essential for \chdeleted{creating} interactive virtual environments with physically plausible behaviors.

Current approaches to this problem fall into two main categories, each with significant limitations. One dominant paradigm is test-time optimization \citep{gradsim, li2023pac, zhong2024springgaus, huang2024dreamphysics, lin2025omniphysgs, zhang2024physdreamer}, which iteratively refines material parameters for each new scene by backpropagating through a differentiable simulator. While powerful, this process is notoriously slow, often taking hours per object, and fails to generalize across scenes. A more recent and promising direction is feed-forward prediction, \chadded{exemplified by} PIXIE~\cite{le2025pixie}. This approach offers fast, generalizable inference by training a network on large-scale datasets. However, this prior method is fundamentally deterministic, yielding only a single point estimate for an object's properties. This overlooks a crucial aspect of physical reality: \textit{ambiguity}. For instance, a visually identical object can possess a spectrum of plausible stiffness. To address this, we focus on modeling the primary axis of this ambiguity: the continuous spectrum from an object’s softest to its stiffest plausible state.

In this paper, we reframe physics prediction not as a deterministic regression task, but as a problem of learning a \textit{controllable physical spectrum}. We introduce \method, a novel framework that learns this continuous path of material parameters. Building upon the feed-forward paradigm of its predecessor PIXIE~\cite{le2025pixie}, which utilized a U-Net to predict a single point estimate, \method introduces a novel conditional framework and a \chdeleted{more powerful} Transformer-based architecture. To facilitate the probabilistic learning problem, we first introduce \dataset, a new large-scale dataset created by augmenting the original PIXIEVERSE \cite{le2025pixie}  
with physically-grounded property ranges, enabling supervised learning of this continuous physical \textit{spectrum}. At its core, \method leverages a unified \textit{Perceiver-IO-like encoder} \cite{jaegle2021perceiver} and a suite of conditional \textit{Flow Matching decoders} \cite{lipman2022flow} to explicitly model the mapping along an object's softest-to-stiffest plausible states. This design enables intuitive and controllable inference: by simply interpolating a single parameter, $\alpha$, users can generate a continuous spectrum of valid material fields, yielding a diverse range of simulation outcomes from a single visual input (see Figure~\ref{fig:hero}).

Furthermore, \method addresses another critical limitation of all prior works: simulator-specific parameterization. Existing feed-forward and test-time optimization methods are often tightly coupled with a single simulation paradigm, such as the Material Point Method (MPM) ~\cite{jiang2016mpm, le2025pixie, xie2024physgaussian}, which limits the portability of the predicted parameters. Our work is the first to propose a \textit{unified architecture} that produces consistent, simulation-ready parameters for fundamentally different downstream physics solvers. To achieve this, we propose to learn implicit physics in a \textit{shared latent space} that can then be decoded via different heads for various physics solvers including MPM, Linear Blend Skinning (LBS) for reduced-order simulators like Simplicits~\cite{modi2024simplicits}, and Spring-Mass systems \cite{zhong2024springgaus}, significantly enhancing the versatility of physics prediction.

In summary, our main contributions include:
\begin{enumerate}
    \item A novel framework, \method, that enables \textit{controllable generation} of a \textit{continuous spectrum} of plausible material properties from a single visual input.
    \item The introduction of \dataset, a new large-scale dataset with material property range annotations to facilitate research in generative physics modeling.
    \item The first \textit{unified multi-solver architecture} capable of producing consistent, simulation-ready parameters for diverse physics backends (MPM, LBS, and Spring-Mass).
\end{enumerate}

Extensive experiments show that \method achieves state-of-the-art accuracy (over 50\% improvement) on our benchmark while enabling a rich spectrum of physically plausible simulations, and retaining fast inference and generalizability advantage of feed-forward prediction.
\vspace{-3pt}
\section{Related Work}
\vspace{-3pt}
\label{sec:related}


\subsection{Inferring Physical Properties from Vision}

The task of estimating an object's physical properties from its appearance is a long-standing challenge in computer vision. Early approaches often relied on analyzing object motion \chreplaced{from video or sensor observations}{in videos} to solve an inverse problem~\cite{davis2015visual, wu2015galileo, wang2015dcms}. More recent methods can be broadly categorized into three paradigms: test-time optimization, VLM-based approaches, and feed-forward networks.

\noindent\textbf{Test-Time Optimization.} A significant body of work leverages differentiable physics simulators to iteratively optimize material parameters. These methods refine a material field by minimizing the discrepancy between a simulated outcome and a ground-truth observation~\cite{gradsim, li2023pac, zhong2024springgaus} or a realism score from a video generation model~\cite{huang2024dreamphysics, lin2025omniphysgs, zhang2024physdreamer}. While capable of producing high-fidelity, scene-specific results, this paradigm suffers from extremely slow, per-scene optimization, often taking hours, \chreplaced{and fails to generalize}{and lacks the ability to generalize} to new objects without re-optimizing from scratch.

\noindent\textbf{VLM-Based Zero-Shot Prediction.} To avoid costly optimization, another line of work directly queries large Vision-Language Models (VLMs) at inference time to obtain physical property estimates. NeRF2Physics~\cite{zhai2024physical} and PUGS~\cite{shuai2025pugszeroshotphysicalunderstanding} associate language-grounded features in NeRFs or Gaussian splats with LLM-generated material dictionaries via retrieval-based regression. While fast and versatile, these approaches are limited by the VLM's inherent knowledge, can produce noisy or inconsistent predictions, and typically only assign coarse, part-level properties rather than fine-grained volumetric fields.

\noindent\textbf{Feed-Forward Supervised Prediction.} Our work belongs to the emerging paradigm of training a generalizable, feed-forward network on large-scale datasets. PIXIE~\cite{le2025pixie}, a prior work, demonstrated the viability of this approach by training a U-Net to predict a single, deterministic set of material properties from distilled CLIP features. \method improves upon this work by predicting a full spectrum of materials via a flow matching model. This generative formulation moves beyond a single point estimate to capture the inherent ambiguity of physical properties from vision, enabling controllable generation of a continuous range of dynamic behaviors.


\subsection{Generative Models for Physics}

The integration of generative models, particularly diffusion and flow-matching models, with physics has recently gained significant attention. Many works focus on generating realistic dynamic videos by incorporating physical priors or constraints into the generation process~\cite{li2024generative, li2025wonderplay}. For instance, WonderPlay~\cite{li2025wonderplay} employs a hybrid simulator that uses a physics engine to generate coarse motion, which then conditions a video diffusion model to produce realistic renderings. PhysCtrl~\cite{physctrl2025} \chdeleted{trains a neural network that }generates point trajectories for different materials with simulation data, achieving simulation-free action to video generation. \chreplaced{Unlike these approaches that produce non-interactive 
pixel sequences, \method generates \textit{simulation-ready, 
reusable} physical parameters deployable in standard 
engines, enabling interactive dynamics rather than 
one-off visual generation.}{ While these methods excel at generating visually plausible \textit{outcomes}, they do not explicitly produce reusable, simulator-agnostic physical \textit{parameters}.}

\chreplaced{Closer to our work are generative models for 3D physics, such as PhysX-3D~\cite{cao2025physx3d}. While unified representations have emerged for decoding static visual formats (\emph{e.g.}, TRELLIS~\cite{xiang2024trellis}), \method is the first to adapt this paradigm for diverse physics solvers. Unlike methods generating entirely new assets, we augment existing 3D objects with a controllable physical continuum. To our knowledge, \method is the first conditional flow-matching framework explicitly generating a continuous range of volumetric material fields for static 3D objects.}{Closer to our work are methods that generate 3D assets along with their physical properties, such as PhysX-3D~\cite{cao2025physx3d}. These approaches typically learn a joint distribution over shape and physics to generate entirely new, simulation-ready assets. Different from these asset generation methods, our goal is to analyze and augment existing 3D objects with a controllable spectrum of physical properties. To our knowledge, \method is the first to employ a conditional flow-matching framework specifically for the task of generating a continuous range of volumetric material fields for a given static 3D object.}

\subsection{Material Point and Reduced-Order Simulation}

Our framework's ability to predict parameters for multiple, fundamentally different physics solvers is a key contribution. Most prior work in physics prediction is tightly coupled to a single simulation paradigm, most commonly the Material Point Method (MPM)~\cite{le2025pixie, xie2024physgaussian}, a continuum-based approach. While powerful, MPM is computationally intensive and not always the ideal choice.

Reduced-order models offer a more efficient alternative by simplifying the deformation space. Vid2Sim~\cite{chen2025vid2sim} \chreplaced{leverages a mesh-free}{demonstrates the effectiveness of a mesh-free} simulator based on Linear Blend Skinning (LBS), which models deformation via a set of learned control handles. Similarly, Spring-Gaus~\cite{zhong2024springgaus} and PhysTwin~\cite{jiang2025phystwin} introduce an efficient simulator based on an anchor-based Spring-Mass system. However, these approaches focus on system identification from video rather than predicting parameters directly from static visual features.

\vspace{-3pt}
\section{Method}
\label{sec:method}
\vspace{-3pt}

\begin{figure*}[t]
    \centering
\includegraphics[width=\textwidth]{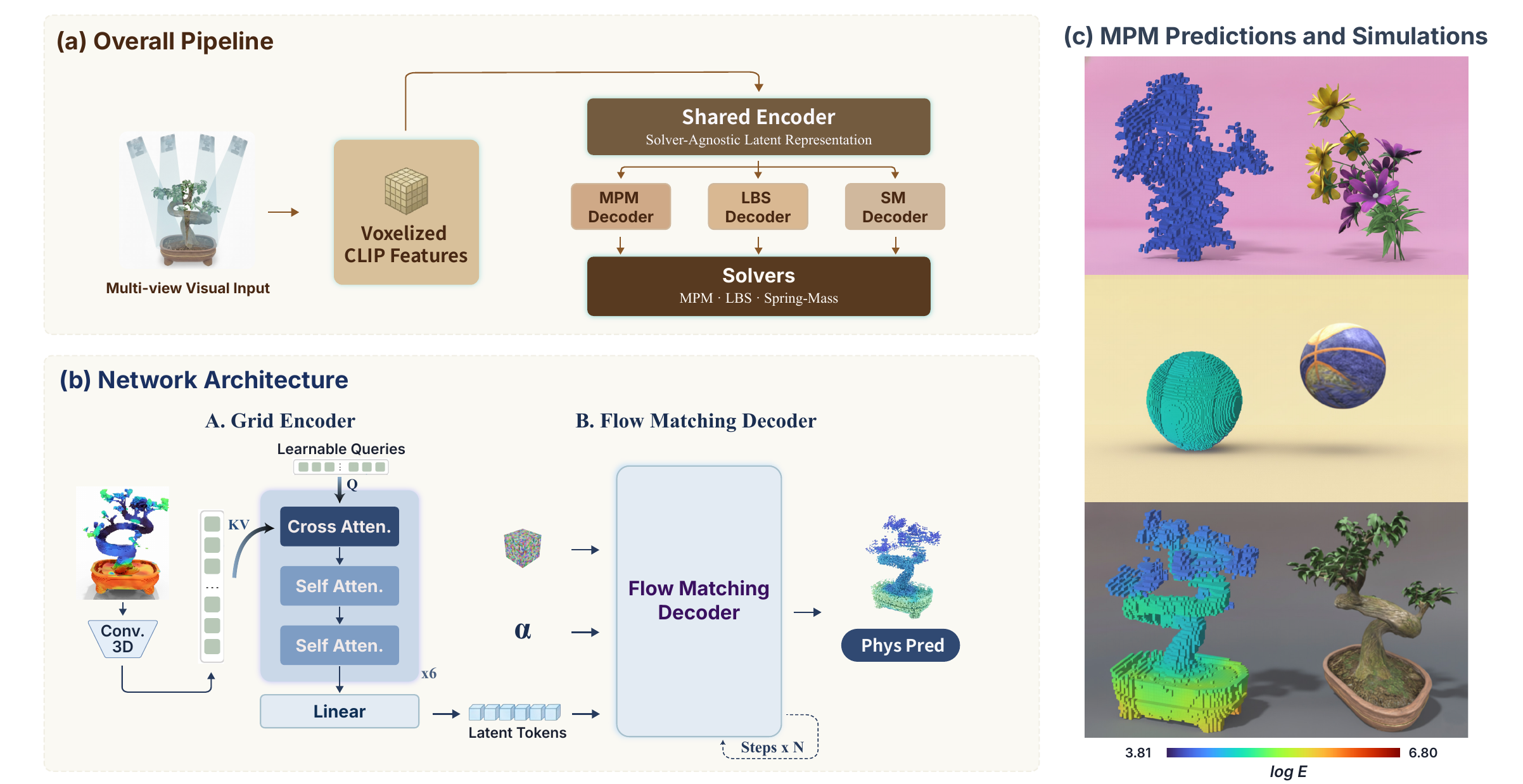}
    \caption{\textbf{Overview of the \method Framework.} Our method generates controllable physical properties from visual input via a unified encoder-decoder architecture.
\textbf{(a) Overall Pipeline:} Multi-view CLIP features are voxelized and processed by the unified encoder. The resulting solver-agnostic latent representation is then passed to three specialized decoders \chreplaced{with a shared architecture but separate parameters,}{, which share the same network architecture} to produce parameters for specific physics engines: Material Point Method (MPM), Linear Blend Skinning (LBS), and Spring-Mass (SM).
\textbf{(b) Network Architecture:} A Grid Encoder distills visual features from a convolutional backbone into latent tokens using a stack of cross-attention and self-attention blocks. A Flow Matching Transformer decoder then uses these tokens and a control parameter $\alpha$ to generate the final physical property fields.
\textbf{(c) MPM Predictions and Simulation:} We show qualitative results for the MPM solver, visualizing the predicted voxelized Young's Modulus field alongside a rendered frame from its physics simulation.}
\label{fig:method_overview}

\end{figure*}

We introduce \method, a feed-forward framework for generating a controllable distribution of physical properties from visual input. Our approach is built on a portable encoder-decoder architecture that generates simulation-ready parameters for diverse physics solvers. The overall pipeline is illustrated in Figure~\ref{fig:method_overview}(a).

\subsection{Physics-Aware Latent Representation}
\label{sec:representation}

\noindent \textbf{Visual Feature Aggregation.} Following PIXIE~\cite{le2025pixie}, we begin by distilling rich visual priors from a pre-trained CLIP model \cite{radford2021learning} into a 3D representation. For a given object, we render multi-view images and extract dense CLIP feature maps. These 2D features are then lifted into 3D space and voxelized, resulting in a sparse feature grid $\mathcal{G}_{\text{feat}} \in \mathbb{R}^{N \times N \times N \times D}$, where $N=64$ is the grid resolution and $D=768$ is the feature dimension. This grid serves as the input to our model, encoding \chdeleted{both} the geometry, appearance, and semantic information of the object.

\noindent \textbf{Latent Encoding.}
To produce a unified latent embedding suitable for multi-solver decoding, our framework employs a Grid Encoder $\mathcal{E}$, whose architecture is shown in Figure~\ref{fig:method_overview} (b). Inspired by Perceiver-IO~\cite{jaegle2021perceiver}, the encoder operates in two stages. First, a 3D convolutional backbone progressively downsamples the input grid from its initial $64^3$ resolution to $16^3$, simultaneously reducing computational cost for the subsequent attention layers and encouraging the extraction of higher-level geometric structure. Second, a tokenizer composed of $N_{\text{blocks}}$ cascaded blocks updates a set of $L$ learnable latent queries: each block first performs cross-attention from the latent queries to the convolutional features, followed by two self-attention layers that refine the latent representation. This yields the final representation:
\begin{equation}
\boldsymbol{z}_{\text{latent}} = \mathcal{E}(\mathcal{G}_{\text{feat}}) \in \mathbb{R}^{L \times C} \enspace,
\end{equation}
where $L$ is the number of latent tokens and $C$ is their dimension. The resulting set of tokens $\boldsymbol{z}_{\text{latent}}$ forms a unified, solver-agnostic latent representation of \chdeleted{the object's} physics-aware geometry, which is key to the framework's portability.

\subsection{Conditional Generation via Flow Matching}
\label{sec:generation}
The physical properties are generated by a conditional Flow Matching Transformer (FMT) decoder, shown in Figure~\ref{fig:method_overview}(b).
To enable controllable generation, we introduce a single, intuitive, scalar parameter $\alpha \in [0, 1]$ that represents the interpolation coefficient from an object's softest ($\alpha=0$) to its stiffest ($\alpha=1$) state. The target physical property $\boldsymbol{y}_{\text{target}}$ for a training instance is generated via Linear Interpolation (LERP):
\begin{equation}
\label{eq:lerp}
\boldsymbol{y}_{\text{target}} = (1-\alpha)\boldsymbol{y}_{\min} + \alpha \boldsymbol{y}_{\max} \enspace.
\end{equation}
This target $\boldsymbol{y}_{\text{target}}$ serves as $\boldsymbol{x}_1$ in the Conditional Flow Matching (CFM) objective~\cite{lipman2022flow}. The model learns a vector field $\boldsymbol{v}_\theta$ that transforms a noise sample $\boldsymbol{x}_0 \sim \mathcal{N}(0, \boldsymbol{I})$ to this target. The CFM loss is:
\begin{equation}
    \mathcal{L}_{\text{CFM}} = \mathbb{E}_{t, \boldsymbol{x}_0, \boldsymbol{y}_{\text{target}}, \boldsymbol{c}} \| \boldsymbol{v}_\theta(\boldsymbol{x}_t, t, \boldsymbol{c}) - (\boldsymbol{y}_{\text{target}} - \boldsymbol{x}_0) \|^2_2 \enspace,
\end{equation}
where $\boldsymbol{x}_t = (1-t)\boldsymbol{x}_0 + t\boldsymbol{y}_{\text{target}}$. The control parameter $\alpha$ is encoded into the conditioning signal $\boldsymbol{c}$ and used to modulate the transformer blocks via adaptive layer normalization (AdaLN), allowing for precise control over the generated physical spectrum.

\begin{figure*}[t]
    \centering
\includegraphics[width=\textwidth]{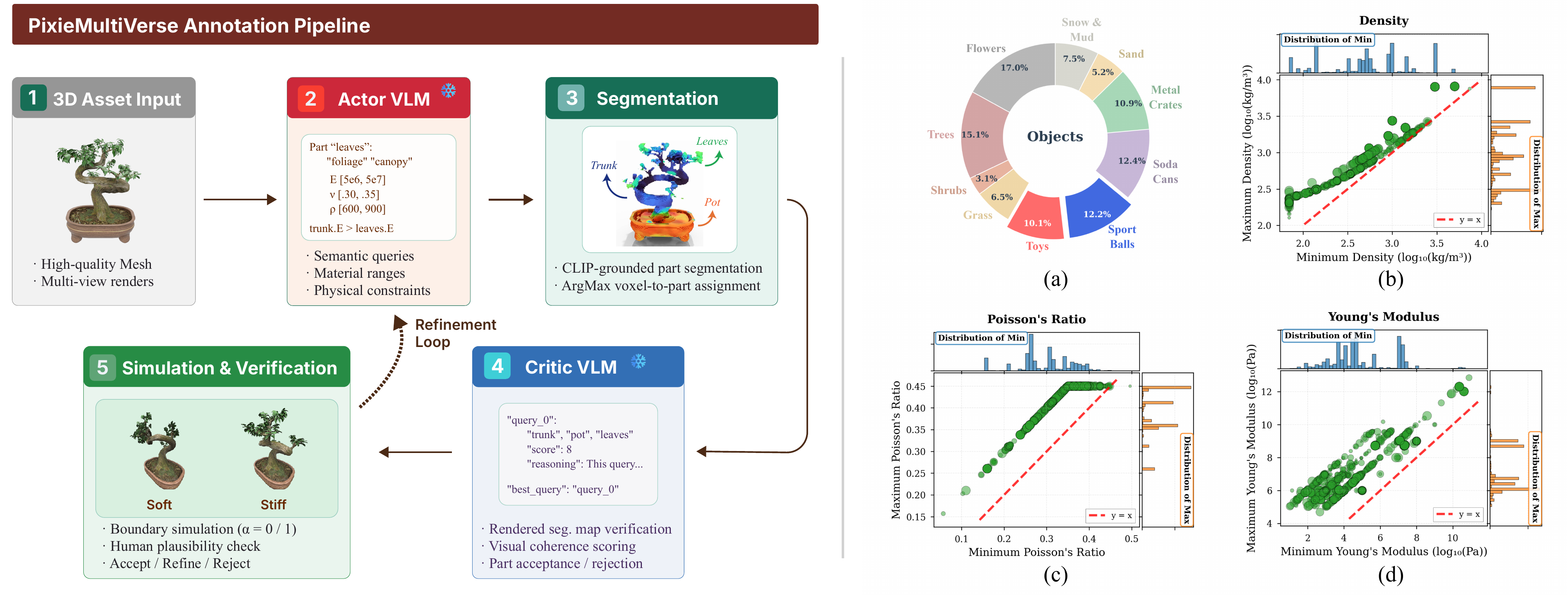}
    \caption{\textbf{\dataset: Annotation Pipeline and Data Overview.} We introduce a dataset \chreplaced{with annotated material property ranges}{with physical property distribution} for controllable generation. Our semi-automatic annotation pipeline \chreplaced{employs an Actor-Critic VLM design with human verification, extending }{, which extends the process from}PIXIE~\cite{le2025pixie}, \chdeleted{is used} to label 10 semantic object classes \textbf{(a)}. We show the resulting distributions of annotated ranges for MPM solver parameters: density \textbf{(b)}, Poisson's ratio \textbf{(c)}, and Young's modulus \textbf{(d)}, which serve as the foundation for our multi-solver framework.}
    \label{fig:dataset_overview}
\end{figure*}

\subsection{Multi-Solver Parameter Decoding}
\label{sec:decoding}
A key innovation of \method is its ability to generate simulation-ready parameters for diverse physics engines through specialized decoders, all conditioned on the same latent tokens $\boldsymbol{z}_{\text{latent}}$, as visualized in our overall pipeline (Figure~\ref{fig:method_overview}, a).

\vspace{4pt}
\noindent\textit{(a) Material Point Method (MPM).} A Flow Matching Transformer decoder $\mathcal{D}_{\text{MPM}}$ generates a spatially-varying material field for all $K$ foreground voxels. The output is a set of physical properties $\mathcal{M}_{\text{MPM}}$:
\begin{equation}
    \mathcal{D}_{\text{MPM}}: (\boldsymbol{z}_{\text{latent}}, \alpha) \rightarrow \mathcal{M}_{\text{MPM}} = \{ (E_i, \nu_i, \rho_i, l_i) \}_{i=1}^{K} \enspace,
\end{equation}
where $E, \nu, \rho$ are the continuous Young's modulus, Poisson's ratio, and density  (respectively), and $l$ is the \chreplaced{categorical material class}{discrete material class ID}.

\vspace{4pt}
\noindent\textit{(b) Linear Blend Skinning (LBS).} \chreplaced{We employ a dual-decoder approach. While the continuous material properties ($E, \nu$) are volumetric fields generated by our standard FMT decoder, the deformation model requires a distinct parameterization. Following Vid2Sim~\cite{chen2025vid2sim}, we use a HyperNetwork to regress the object-specific LBS parameters $\theta_{\text{LBS}}$ from the global latent tokens. Crucially, this deformation structure remains static, while the soft-to-stiff continuum is driven entirely by the $\alpha$-conditioned material fields.}{We employ a dual-decoder approach. For LBS, the parameters are structurally different. The material properties (E, $\nu$) are volumetric fields similar to MPM, making them suitable for a FMT decoder. However, the skinning weights are defined by a separate network. Following prior work Vid2Sim~\cite{chen2025vid2sim}, we find it more effective to predict the parameters $\theta_{\text{LBS}}$ of this weight network directly using a HyperNetwork, which is well-suited for regressing a small, structured set of parameters from unified latent tokens.}

\vspace{4pt}
\noindent\textit{(c) Spring-Mass System.} Another Transformer decoder $\mathcal{D}_{\text{Spring}}$ generates a single high-dimensional vector $\boldsymbol{m}_{\text{spring}}$ containing all necessary physics parameters. This adapts the simplified design from Spring-Gaus~\cite{zhong2024springgaus}:
\begin{equation}
    \mathcal{D}_{\text{Spring}}: (\boldsymbol{z}_{\text{latent}}, \alpha) \rightarrow \boldsymbol{m}_{\text{spring}} = (\boldsymbol{k}, \eta) \enspace,
\end{equation}
where $\boldsymbol{k} \in \mathbb{R}^{N_a}$ is the stiffness vector for $N_a$ anchors and $\eta$ is a scalar global softness parameter.

\subsection{Physics Simulation}
\label{sec:physics_simulation}

The parameters generated by \method are designed to be directly consumed by downstream physics solvers. We demonstrate our method's portability across three distinct simulation paradigms.

\noindent\textbf{Material Point Method (MPM).}
\chreplaced{We adopt PhysGaussian~\cite{xie2024physgaussian}, an MPM-based solver that excels at modeling large deformations and contact. Following PIXIE~\cite{le2025pixie}, the solver treats 3D Gaussians as physical particles, each augmented with predicted material properties, to simulate their dynamics under external forces.}{We use MPM for continuum-based simulation, which excels at modeling large deformations and contact. Following PIXIE~\cite{le2025pixie}, the MPM solver takes a point cloud of initial particle positions, each augmented with predicted material properties, and simulates their dynamics under external forces.}

\noindent\textbf{Linear Blend Skinning (LBS).}
For efficient, reduced-order deformation, we employ an LBS-based simulator inspired by Vid2Sim~\cite{chen2025vid2sim} and Simplicits~\cite{modi2024simplicits}. \chreplaced{To accelerate implicit integration during simulation, a sparse set of cubature points is sampled via Farthest Point Sampling (FPS) from the foreground voxels. Meanwhile, the skinning MLP (parameterized by our model) assigns continuous influence weights to the entire voxelized field relative to a set of control handles, enabling stable and efficient elastic dynamics.}{This approach represents an object's deformation as a weighted combination of transformations applied to a small set of control handles. Our model predicts the parameters for this system, enabling stable elastic dynamics.}

\noindent\textbf{Spring-Mass System.}
To model objects with anchor-based dynamics, we utilize the Spring-Mass solver from SpringGaus~\cite{zhong2024springgaus}. This system represents an object as a network of interconnected mass points and springs. Our model predicts the stiffness of the springs and a global softness parameter, allowing for control over the object's overall compliance.

\begin{table*}[t]
  \centering
  \small
  \caption{\textbf{Quantitative Comparison of Physical Property Regression.} \method sets a new state-of-the-art in continuous property prediction, reducing Young's Modulus MSE by over 50\% compared to the specialized PIXIE. We compare our generative models (averaged across $\alpha \in \{0.0, 0.5, 1.0\}$) against deterministic baselines (single-point predictions). While PIXIE retains a slight edge in discrete material classification, \method excels across all continuous physical parameters. The $\pm$ denotes standard deviation (95\% CI for Accuracy). * indicates baselines adapted or re-trained on our dataset for fair comparison. Best results are \textbf{bolded}; second-best are \underline{underlined}. \chadded{Runtime for \method reflects MPM-only single-decoder 
inference; full three-solver generation takes 
$\sim$21.6s (see Table~\ref{tab:solver_clustered_comparison})}.
  }
  \label{tab:simple_metrics_comparison}
  \resizebox{\textwidth}{!}{%
  \begin{tabular}{l c c c c c c c}
    \toprule
    \textbf{Method} 
    & \textbf{SSIM} ($\uparrow$) 
    & \textbf{PSNR} ($\uparrow$) 
    & \textbf{$\log E$ MSE} ($\downarrow$)
    & \textbf{$\log \rho$ MSE} ($\downarrow$)
    & \textbf{$\nu$ MSE} ($\downarrow$)
    & \textbf{Material Acc.} ($\uparrow$)
    & \textbf{Runtime} ($\downarrow$) \\
    \midrule
    \multicolumn{8}{l}{\textit{Deterministic Baselines}} \\
    \midrule
    
    NeRF2Physics~\cite{zhai2024physical} 
    & 0.879 {\small\textcolor{gray}{±0.095}}
    & 24.84 {\small\textcolor{gray}{±9.45}}
    & 0.5236 {\small\textcolor{gray}{±0.61}}
    & 0.2958 {\small\textcolor{gray}{±0.36}}
    & 0.3430 {\small\textcolor{gray}{±0.38}}
    & 63.4\% {\small\textcolor{gray}{±6.55\%}}
    & 119.49s {\small\textcolor{gray}{±31.28}} \\
    
    PUGS*~\cite{shuai2025pugszeroshotphysicalunderstanding}
    & 0.886 {\small\textcolor{gray}{±0.096}}
    & 24.21 {\small\textcolor{gray}{±8.77}}
    & 1.0591 {\small\textcolor{gray}{±0.60}}
    & 0.2335 {\small\textcolor{gray}{±0.36}}
    & --- 
    & --- 
    & 36.31s {\small\textcolor{gray}{±15.79}} \\
    
    PIXIE*~\cite{le2025pixie}
    & \textbf{0.922} {\small\textcolor{gray}{±0.085}}
    & 30.17 {\small\textcolor{gray}{±11.50}}
    & \underline{0.0205} {\small\textcolor{gray}{±0.06}}
    & \underline{0.0244} {\small\textcolor{gray}{±0.06}}
    & \underline{0.0557} {\small\textcolor{gray}{±0.12}}
    & \textbf{97.3\%} {\small\textcolor{gray}{±0.99\%}}
    & \textbf{0.137s} {\small\textcolor{gray}{±0.03}} \\
    
    \midrule
    \multicolumn{8}{l}{\textit{Our Generative Models (avg across 3 $\alpha$)}} \\
    \midrule
    
    3D U-Net (Ablation)
    & 0.909 {\small\textcolor{gray}{±0.09}}
    &  \underline{30.75} {\small\textcolor{gray}{±10.86}}
    & 0.0410 {\small\textcolor{gray}{±0.08}}
    & 0.1464 {\small\textcolor{gray}{±0.16}}
    & 0.4604 {\small\textcolor{gray}{±0.48}}
    & \underline{96.3\%} {\small\textcolor{gray}{±1.0\%}}
    &  \underline{10.77s}{\small\textcolor{gray}{±1.20}} \\
    
    \textbf{\method (Ours)}
    & \underline{0.916} {\small\textcolor{gray}{±0.03}}
    & \textbf{30.83} {\small\textcolor{gray}{±12.82}}
    & \textbf{0.0091} {\small\textcolor{gray}{±0.03}}
    & \textbf{0.0194} {\small\textcolor{gray}{±0.004}}
    & \textbf{0.0240} {\small\textcolor{gray}{±0.05}}
    & 93.9\% {\small\textcolor{gray}{±1.0\%}}
    & 12.16s{\small\textcolor{gray}{±0.01}} \\
    
    \bottomrule
  \end{tabular}
  }
\end{table*}

\subsection{\dataset Dataset}
\label{sec:dataset}
We introduce \dataset, a large-scale dataset of 3D objects annotated with diverse physical property \textit{ranges} to facilitate controllable generation. Our dataset is built upon the 1410 high-quality assets from PIXIEVERSE~\cite{le2025pixie}, which we re-annotate using a semi-automatic pipeline illustrated in Figure~\ref{fig:dataset_overview}.

\noindent\textbf{MPM Range Annotation.} While extending PIXIE's Actor-Critic segmentation pipeline, our core contribution lies in annotating a continuous spectrum of plausible material properties rather than deterministic point values. Specifically, we prompt \textbf{GPT-4o} (Actor) to explicitly propose bounding ranges $[\boldsymbol{y}_{\min}, \boldsymbol{y}_{\max}]$ for physical parameters alongside crucial inter-part constraints (\emph{e.g.}, ensuring $E_\text{trunk} \gg E_\text{leaf}$). \textbf{Gemini-2.5-Flash} (Critic) then selects the most geometrically coherent query set by evaluating rendered 3D segmentation maps, where each voxel is assigned to a part via \textit{argmax} over CLIP~\cite{radford2021learning} feature similarities. Crucially, to guarantee physical plausibility and visual diversity, these VLM-proposed ranges are manually verified and refined through boundary-value simulations (at $\alpha=0$ and $\alpha=1$). A rigorous quality audit confirmed an \textbf{8.9\%} rejection rate and a \textbf{38.3\%} refinement frequency (details in the supplement). The resulting diverse MPM parameter distributions are shown in Figure~\ref{fig:dataset_overview}~(b-d).

\noindent\textbf{Cross-Solver Parameter Generation.} \chreplaced{For solvers like LBS and Spring-Mass systems with distinct parameter sets, we bypass direct annotation. Instead, we generate ground-truth MPM simulations exclusively at the soft ($\alpha=0$) and stiff ($\alpha=1$) boundaries. Slow test-time methods~\cite{chen2025vid2sim, zhong2024springgaus} are then applied to fit solver-specific parameters to these boundary videos, followed by manual verification and refinement to correct fitting inaccuracies. During training, target labels for any intermediate $\alpha$ are derived via interpolation. This efficiently ensures that target dynamic behaviors remain consistent across all solvers.}{For solvers like Linear Blend Skinning (LBS) and Spring-Mass systems, which operate on their own distinct parameter sets, we do not perform direct annotation. Instead, we leverage slow test-time methods \cite{chen2025vid2sim, zhong2024springgaus} to obtain the labels, which will be distilled back to our feed-forward model. To obtain the ground-truth video needed for test-time optimization, we generate a video simulated via MPM using a randomly sampled $\alpha$. This ensures that the target dynamic behavior for a given control state $\alpha$ is consistent across all solvers, a critical property for training our unified model.}

\section{Experiments}
\label{sec:experiments}
We conduct a comprehensive set of experiments to validate the three core contributions of \method. We first establish its accuracy against state-of-the-art deterministic methods (\S\ref{sec:exp_accuracy}). We then evaluate the effectiveness and portability of our novel unified multi-solver architecture by comparing it against specialized, single-task models (\S\ref{sec:exp_multisolver}). Finally, we analyze its primary generative capability: the controllable generation of a continuous physical spectrum (\S\ref{sec:exp_generative}).

\subsection{Experimental Setup}

\paragraph{Baselines.}
We compare \method against several baselines. For a fair comparison, we adapted all baseline methods to our task. For NeRF2Physics~\cite{zhai2024physical} and PUGS~\cite{shuai2025pugszeroshotphysicalunderstanding}, we queried its VLM using prompts derived from our object classes. For PIXIE~\cite{le2025pixie}, 
we conducted a full re-training on our \dataset dataset, using the midpoint ($\alpha=0.5$) properties as single ground truth target. As key ablation, we also train a 3D diffusion U-Net as a baseline generative model. Detailed training hyper-parameters for baselines are provided in the appendix.

\paragraph{Dataset and Metrics.}

All models are trained on the proposed \textbf{\dataset} dataset. Evaluation is conducted on a hold-out test set. For the MPM solver, the full test split consists of 41 objects. For the LBS and Spring-Mass solvers, which are intended for elastic materials, we evaluate on a 10-object subset corresponding to this category (for example, \textit{Toys} and \textit{Sport Balls}). To evaluate \textit{property prediction accuracy}, we report the Mean Squared Error (MSE) for continuous properties in log-space ($\log E, \log \rho$) and linear-space ($\nu$), alongside material classification Accuracy. To evaluate simulation quality, we measure the fidelity of video reconstruction using PSNR, SSIM~\cite{wang2004ssim}, and LPIPS~\cite{zhang2018unreasonable} against ground-truth simulations.

\subsection{Accuracy against Deterministic Baselines}
\label{sec:exp_accuracy}

We first validate that our generative framework can produce single-point estimates that are more accurate than prior deterministic methods. For a fair comparison, we evaluate \method's average performance across its distribution against the single-point predictions of baselines.

\noindent\textbf{Quantitative Analysis.} Table~\ref{tab:simple_metrics_comparison} presents the results. \method sets a new state of the art in physical property regression. It achieves an MSE of \textbf{0.0091} for Young's Modulus ($\log E$), which is more than twice as accurate as the previous best method, PIXIE. \chreplaced{We further compare against a VLM-feedback baseline (Gemini-3.0-Flash + iterative MPM loop), finding \method is 128$\times$ faster with significantly lower $\log E$ MSE (details in the supplement).}{This result is significant: it shows that by learning a continuous distribution, our model develops a more robust and accurate underlying representation than models trained on a single point estimate.}

\begin{figure*}[t]
    \centering
\includegraphics[width=\textwidth]{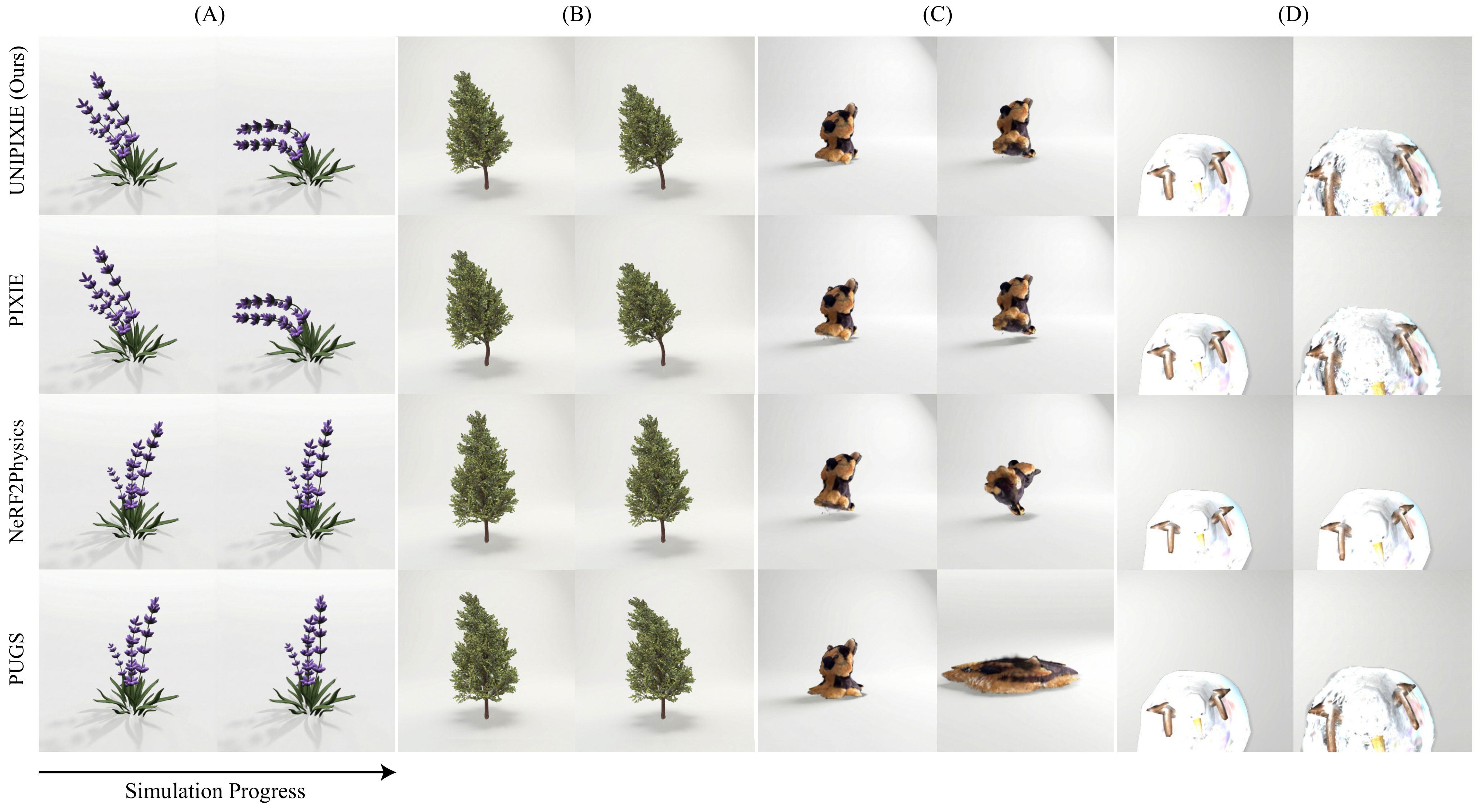}
\caption{\textbf{Qualitative Comparison of Predicted Dynamics.} When evaluated at its midpoint ($\alpha=0.5$), our model generates physically plausible simulations competitive with the specialist PIXIE and avoid the failure modes of other baselines. This figure compares a mid-simulation frame (left) and the final state (right) for each method. We observe that NeRF2Physics and PUGS often produce unnaturally rigid motion for flexible objects like the lavender plant (A) and tree (B), a result of predicting an overly high Young's modulus. Furthermore, PUGS can suffer from critical material misclassification, causing the teddy bear (C) to unrealistically collapse. While baselines can occasionally yield plausible dynamics for specific materials (e.g., the collapsing snow in D), they fail to generalize. In contrast, our model avoids these pitfalls and demonstrates robust physical understanding across all diverse scenarios.}
\label{fig:qualitative_comparison}
\end{figure*}

Figure~\ref{fig:qualitative_comparison} \chreplaced{qualitatively compares }{provides a qualitative comparison of }the resulting simulations (evaluated at $\alpha=0.5$ for fair comparison). The visualizations confirm the quantitative findings, showing that \method consistently generates stable and plausible dynamics, avoiding common failure modes like the unnatural rigidity of NeRF2Physics or the simulation collapse of PUGS.


\begin{table*}[t]
  \centering
  \small
  \caption{\textbf{Solver-specific Quantitative Comparison.} Our single unified \method model achieves performance competitive with or superior to specialized state-of-the-art methods across diverse physics solvers, while being orders of magnitude faster than test-time optimization baselines. We evaluate video reconstruction fidelity (PSNR, SSIM, LPIPS) across the full physical distribution: soft ($\alpha=0.0$), mid ($\alpha=0.5$), and stiff ($\alpha=1.0$). The $\pm$ denotes standard deviation. Best results within each solver category are \textbf{bolded}.
  }
    \label{tab:solver_clustered_comparison}
  \setlength{\tabcolsep}{5pt} 
  \resizebox{\textwidth}{!}{%
  \begin{tabular}{l ccc ccc ccc c}
    \toprule
    \multirow{2}{*}{\textbf{Method}} & \multicolumn{3}{c}{\textbf{$\alpha = 0.0$ (Soft)}} & \multicolumn{3}{c}{\textbf{$\alpha = 0.5$ (Mid)}} & \multicolumn{3}{c}{\textbf{$\alpha = 1.0$ (Stiff)}} &
    \multirow{2}{*}{\textbf{Runtime (s)} $\downarrow$} \\
    \cmidrule(lr){2-4} \cmidrule(lr){5-7} \cmidrule(lr){8-10}
    & \textbf{PSNR ($\uparrow$)} & \textbf{SSIM ($\uparrow$)} & \textbf{LPIPS ($\downarrow$)} & \textbf{PSNR ($\uparrow$)} & \textbf{SSIM ($\uparrow$)} & \textbf{LPIPS ($\downarrow$)} & \textbf{PSNR ($\uparrow$)} & \textbf{SSIM ($\uparrow$)} & \textbf{LPIPS ($\downarrow$)} \\
    \midrule
    \multicolumn{11}{l}{\textit{Solver: Material Point Method (MPM)}} \\
    
    PIXIE~\cite{le2025pixie}
    & 23.16 {\scriptsize\textcolor{gray}{±8.82}} 
    & 0.8868 {\scriptsize\textcolor{gray}{±0.1007}} 
    & 0.1156 {\scriptsize\textcolor{gray}{±0.0969}} 
    & 30.17 {\scriptsize\textcolor{gray}{±11.50}} 
    & \textbf{0.9225} {\scriptsize\textcolor{gray}{±0.0845}} 
    & \textbf{0.0542} {\scriptsize\textcolor{gray}{±0.0697}} 
    & 26.04 {\scriptsize\textcolor{gray}{±9.79}} 
    & 0.9036 {\scriptsize\textcolor{gray}{±0.0929}} 
    & 0.0928 {\scriptsize\textcolor{gray}{±0.0940}} 
    & \textbf{0.14} {\scriptsize\textcolor{gray}{±0.03}} \\
    
    \textbf{\method (Ours)} 
    & \textbf{29.25} {\scriptsize\textcolor{gray}{±12.97}} 
    & \textbf{0.9050} {\scriptsize\textcolor{gray}{±0.1100}} 
    & \textbf{0.1122} {\scriptsize\textcolor{gray}{±0.1143}} 
    & \textbf{30.43} {\scriptsize\textcolor{gray}{±11.98}} 
    & 0.9198 {\scriptsize\textcolor{gray}{±0.0929}} 
    & 0.0810 {\scriptsize\textcolor{gray}{±0.0898}} 
    & \textbf{32.87} {\scriptsize\textcolor{gray}{±13.23}} 
    & \textbf{0.9246} {\scriptsize\textcolor{gray}{±0.0950}} 
    & \textbf{0.0915} {\scriptsize\textcolor{gray}{±0.1021}} 
    & 21.64 {\scriptsize\textcolor{gray}{±0.23}} \\
    
    \midrule
    \multicolumn{11}{l}{\textit{Solver: Linear Blend Skinning (LBS)}} \\
    
    Vid2Sim (full)~\cite{chen2025vid2sim}
    & 28.30 {\scriptsize\textcolor{gray}{±4.66}} 
    & 0.9583 {\scriptsize\textcolor{gray}{±0.0192}} 
    & 0.0595 {\scriptsize\textcolor{gray}{±0.0497}} 
    & \textbf{36.94} {\scriptsize\textcolor{gray}{±5.44}} 
    & 0.9818 {\scriptsize\textcolor{gray}{±0.0142}} 
    & 0.0121 {\scriptsize\textcolor{gray}{±0.0133}} 
    & 40.13 {\scriptsize\textcolor{gray}{±7.94}} 
    & 0.9858 {\scriptsize\textcolor{gray}{±0.0127}} 
    & 0.0080 {\scriptsize\textcolor{gray}{±0.0080}} 
    & 521.32 {\scriptsize\textcolor{gray}{±47.37}} \\

    Vid2Sim (fast)~\cite{chen2025vid2sim}
    & 27.43 {\scriptsize\textcolor{gray}{±3.06}} 
    & 0.9597 {\scriptsize\textcolor{gray}{±0.0150}} 
    & 0.0469 {\scriptsize\textcolor{gray}{±0.0445}} 
    & 36.39 {\scriptsize\textcolor{gray}{±5.78}} 
    & 0.9791 {\scriptsize\textcolor{gray}{±0.0178}} 
    & 0.0112 {\scriptsize\textcolor{gray}{±0.0151}} 
    & 35.03 {\scriptsize\textcolor{gray}{±3.71}} 
    & 0.9800 {\scriptsize\textcolor{gray}{±0.0136}} 
    & 0.0092 {\scriptsize\textcolor{gray}{±0.0080}} 
    & 139.50 {\scriptsize\textcolor{gray}{±3.81}} \\
    
    \textbf{\method (Ours)} 
    & \textbf{33.83} {\scriptsize\textcolor{gray}{±4.12}} 
    & \textbf{0.9774} {\scriptsize\textcolor{gray}{±0.0105}} 
    & \textbf{0.0206} {\scriptsize\textcolor{gray}{±0.0182}} 
    & 36.81 {\scriptsize\textcolor{gray}{±3.30}} 
    & \textbf{0.9859} {\scriptsize\textcolor{gray}{±0.0061}} 
    & \textbf{0.0071} {\scriptsize\textcolor{gray}{±0.0039}} 
    & \textbf{41.63} {\scriptsize\textcolor{gray}{±6.79}} 
    & \textbf{0.9904} {\scriptsize\textcolor{gray}{±0.0081}} 
    & \textbf{0.0050} {\scriptsize\textcolor{gray}{±0.0049}} 
    & \textbf{21.64} {\scriptsize\textcolor{gray}{±0.23}} \\
    
    \midrule
    \multicolumn{11}{l}{\textit{Solver: Spring-Mass (via Spring-Gaus)}} \\

    Spring-Gaus (tuned)~\cite{zhong2024springgaus}
    & 30.53 {\scriptsize\textcolor{gray}{±1.06}} 
    & 0.9396 {\scriptsize\textcolor{gray}{±0.0089}} 
    & 0.1030 {\scriptsize\textcolor{gray}{±0.0188}} 
    & 37.60 {\scriptsize\textcolor{gray}{±3.03}} 
    & 0.9666 {\scriptsize\textcolor{gray}{±0.0089}} 
    & 0.0326 {\scriptsize\textcolor{gray}{±0.0179}} 
    & 36.57 {\scriptsize\textcolor{gray}{±3.06}} 
    & 0.9636 {\scriptsize\textcolor{gray}{±0.0091}} 
    & 0.0386 {\scriptsize\textcolor{gray}{±0.0223}} 
    & 4375.30 {\scriptsize\textcolor{gray}{±533.67}} \\ 
    
    Spring-Gaus~\cite{zhong2024springgaus}
    & 26.77 {\scriptsize\textcolor{gray}{±4.97}} 
    & 0.9401 {\scriptsize\textcolor{gray}{±0.0134}} 
    & 0.0971 {\scriptsize\textcolor{gray}{±0.0235}} 
    & 24.80 {\scriptsize\textcolor{gray}{±2.50}} 
    & 0.9520 {\scriptsize\textcolor{gray}{±0.0163}} 
    & 0.0634 {\scriptsize\textcolor{gray}{±0.0306}} 
    & 24.38 {\scriptsize\textcolor{gray}{±2.65}} 
    & 0.9496 {\scriptsize\textcolor{gray}{±0.0168}} 
    & 0.0686 {\scriptsize\textcolor{gray}{±0.0327}} 
    & 4375.30 {\scriptsize\textcolor{gray}{±533.67}} \\
    
    \textbf{\method (Ours)} 
    & \textbf{33.88} {\scriptsize\textcolor{gray}{±4.56}} 
    & \textbf{0.9544} {\scriptsize\textcolor{gray}{±0.0169}} 
    & \textbf{0.0688} {\scriptsize\textcolor{gray}{±0.0347}} 
    & \textbf{38.79} {\scriptsize\textcolor{gray}{±4.52}} 
    & \textbf{0.9699} {\scriptsize\textcolor{gray}{±0.0133}} 
    & \textbf{0.0307} {\scriptsize\textcolor{gray}{±0.0211}} 
    & \textbf{38.51} {\scriptsize\textcolor{gray}{±4.03}} 
    & \textbf{0.9696} {\scriptsize\textcolor{gray}{±0.0102}} 
    & \textbf{0.0318} {\scriptsize\textcolor{gray}{±0.0204}} 
    & \textbf{21.64} {\scriptsize\textcolor{gray}{±0.23}} \\
    
    \bottomrule
  \end{tabular}
  }
\end{table*}

\begin{figure*}[t]
    \centering
\includegraphics[width=\textwidth]{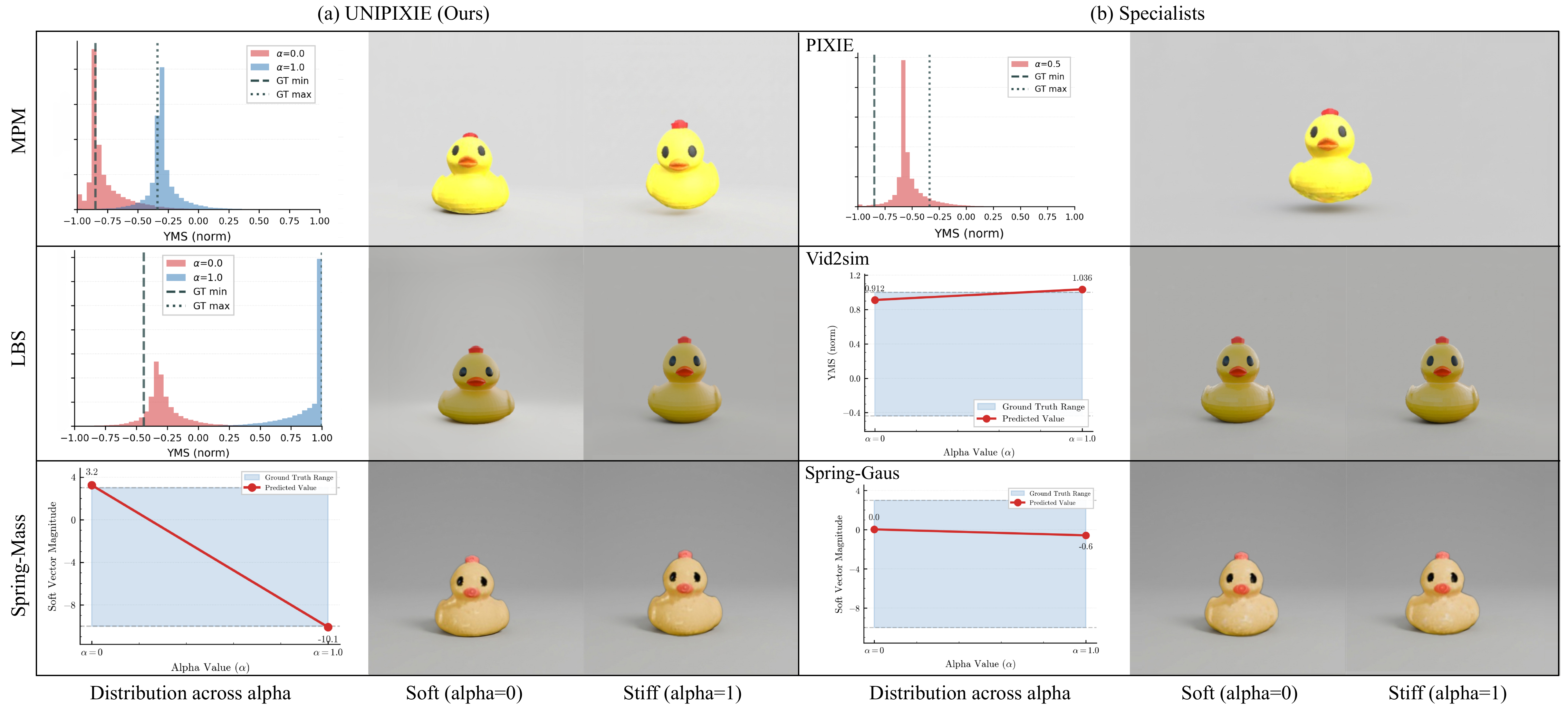}
    \caption{\textbf{Controllable Multi-Solver Generation vs. Specialists.} (a) \method (Ours): Our model learns a smooth soft-to-stiff mapping for diverse solvers, resulting in intuitive deformation changes. (b) Specialists: The simulation quality from our single unified model is comparable to that of \chreplaced{three solver-specific baselines (PIXIE, Vid2Sim, Spring-Gaus)}{three separate specialist models}, confirming its portability and effectiveness.}
    \label{fig:multi_solver_visualization}
\end{figure*}

\subsection{Unified Multi-Solver Evaluation}
\label{sec:exp_multisolver}
Having established its accuracy, we now evaluate the portability and effectiveness of our unified architecture. Table~\ref{tab:solver_clustered_comparison} provides a comprehensive comparison of our single, unified model against specialized models for each physics solver.

\noindent\textbf{Effectiveness.} Our unified model demonstrates remarkable effectiveness, achieving performance that is not just competitive with, but often superior to, specialized models across all three paradigms. For instance, in the LBS setting, our model consistently outperforms both \chadded{the default }Vid2Sim (full) and \chreplaced{its accelerated variant, Vid2Sim (fast), }{the accelerated Vid2Sim (fast) variants,}which uses only one-third of the optimization \chreplaced{steps to speed up inference}{loops for faster inference}. While Vid2Sim (full) narrowly matches our PSNR at the midpoint ($\alpha=0.5$), our model is superior across all other metrics and stiffness levels, particularly in the more challenging soft ($\alpha=0.0$) and stiff ($\alpha=1.0$) regimes. Similarly, for Spring-Mass systems, our model significantly surpasses both the original Spring-Gaus and the stronger Spring-Gaus (tuned) baseline, which benefits from dataset-specific hyperparameter tuning \chadded{to ensure a fair comparison}. The consistent, state-of-the-art performance across disparate solvers proves that our shared encoder learns a powerful latent representation that can be effectively decoded for multiple simulation backends.

\noindent\textbf{Efficiency.} A key advantage of our unified approach is its efficiency. As shown in Table~\ref{tab:solver_clustered_comparison}, \chreplaced{our single feed-forward model generates parameters for all three solvers simultaneously in approximately 21 seconds (12s for MPM alone).}{our single feed-forward model generates parameters in approximately 21 seconds.} This is orders of magnitude faster than the test-time optimization required by Vid2Sim (full) at 521s and Spring-Gaus at 4375s. This significant speed-up, combined with its state-of-the-art effectiveness, makes our method vastly more practical for real-world applications requiring fast, on-demand physics generation.

\subsection{Analysis of Controllable Generation}
\label{sec:exp_generative}

The core novelty of \method is its ability to generate a controllable distribution of physical properties. We validate this capability in Figure~\ref{fig:multi_solver_visualization}.

\noindent\textbf{Distribution and Range Analysis.} The plots in Figure~\ref{fig:multi_solver_visualization}(a) show the distribution of predicted physical parameters for our model at different $\alpha$ settings. The distributions for $\alpha=0.0$ (soft) and $\alpha=1.0$ (stiff) are clearly distinct and correctly aligned with the ground truth min/max boundaries for all three solvers. This confirms that our model has learned to map the control parameter $\alpha$ to a meaningful and well-behaved distribution of physical values.

\noindent\textbf{Qualitative Dynamics.} The corresponding simulation results for the rubber duck show a clear and intuitive change in behavior: at $\alpha=0.0$, the duck is soft and deforms upon impact, while at $\alpha=1.0$, it behaves as a stiff, nearly rigid object. This demonstrates that the learned property spectrum translates directly into a visually diverse and controllable range of dynamic outcomes.

\section{Conclusion}
\label{sec:conclusion}

We introduced \method, a novel framework that reframes physics-from-vision from a deterministic point-estimate task to one of controllable, generative modeling. By learning a continuous soft-to-stiff continuum of material properties on our new \dataset with property range annotations, our model captures a key axis of physical ambiguity. Our experiments demonstrate that this approach not only enables the generation of a diverse range of plausible dynamics but also achieves state-of-the-art prediction accuracy. Furthermore, our unified architecture is the first to produce consistent, simulation-ready parameters for disparate physics backends, significantly enhancing portability. \chreplaced{While our work is a significant step towards more flexible physics prediction, future research could address estimating properties for occluded regions and explore multi-dimensional material manifolds.}{While our work is a significant step towards more flexible physics prediction, future research could explore more complex, multi-dimensional material manifolds.} Nonetheless, \method provides a strong foundation for controllable physical modeling and bridges the gap between static visual perception and the dynamic nature of physical reality.

\section*{Acknowledgment}
This work was partially supported by DARPA grant \#HR00112420305. Any opinions, findings, and conclusions or recommendations expressed in this material are those of the author(s) and do not necessarily reflect the views, position, or policy of DARPA or the US Government.

{
    \small
    \bibliographystyle{ieeenat_fullname}
    \bibliography{main} 

@String(AAAI = {AAAI})

@article{kerbl20233d,
author = {Kerbl, Bernhard and Kopanas, Georgios and Leimkuehler, Thomas and Drettakis, George},
title = {3D Gaussian Splatting for Real-Time Radiance Field Rendering},
year = {2023},
issue_date = {August 2023},
publisher = {Association for Computing Machinery},
volume = {42},
number = {4},
issn = {0730-0301},
url = {https://doi.org/10.1145/3592433},
doi = {10.1145/3592433},
journal = {ACM Trans. Graph.},
month = jul,
articleno = {139}
}

@article{le2025pixie,
  title={Pixie: Fast and Generalizable Supervised Learning of 3D Physics from Pixels},
  author={Long Le and Ryan Lucas and Chen Wang and Chuhao Chen and Dinesh Jayaraman and Eric Eaton and Lingjie Liu},
  journal={arXiv preprint arXiv:2508.17437},
  year={2025},
}

@inproceedings{huang2024dreamphysics,
author = {Huang, Tianyu and Zhang, Haoze and Zeng, Yihan and Zhang, Zhilu and Li, Hui and Zuo, Wangmeng and Lau, Rynson W. H.},
title = {DreamPhysics: learning physics-based 3D dynamics with video diffusion priors},
year = {2025},
url = {https://doi.org/10.1609/aaai.v39i4.32389},
doi = {10.1609/aaai.v39i4.32389},
booktitle = {Proceedings of the AAAI Conference on Artificial Intelligence},
articleno = {416}
}

@inproceedings{lin2025omniphysgs,
  author       = {Yuchen Lin and
                  Chenguo Lin and
                  Jianjin Xu and
                  Yadong Mu},
  title        = {OmniPhysGS: 3D Constitutive Gaussians for General Physics-Based Dynamics Generation},
  booktitle    = {International Conference on Learning Representations},
  year         = {2025},
  url          = {https://openreview.net/forum?id=9HZtP6I5lv}
}

@INPROCEEDINGS{zhai2024physical,
  author={Zhai, Albert J. and Shen, Yuan and Chen, Emily Y. and Wang, Gloria X. and Wang, Xinlei and Wang, Sheng and Guan, Kaiyu and Wang, Shenlong},
  booktitle={Proceedings of the IEEE/CVF Conference on Computer Vision and Pattern Recognition}, 
  title={Physical Property Understanding from Language-Embedded Feature Fields}, 
  year={2024},
  volume={},
  number={},
  pages={28296-28305}
}

@inproceedings{davis2015visual,
  author={Davis, Abe and Bouman, Katherine L. and Chen, Justin G. and Rubinstein, Michael and Durand, Fr{\'{e}}do and Freeman, William T.},
  booktitle={Proceedings of the IEEE/CVF Conference on Computer Vision and Pattern Recognition}, 
  title={Visual vibrometry: Estimating material properties from small motions in video}, 
  year={2015},
  volume={},
  number={},
  pages={5335--5343}
}

@inproceedings{wu2015galileo,
  author       = {Jiajun Wu and
                  Ilker Yildirim and
                  Joseph J. Lim and
                  Bill Freeman and
                  Joshua B. Tenenbaum},
  title        = {Galileo: Perceiving Physical Object Properties by Integrating a Physics
                  Engine with Deep Learning},
  booktitle    = {Advances in neural information processing systems},
  pages        = {127--135},
  year         = {2015}
}

@inproceedings{gradsim,
  author       = {J. Krishna Murthy and
                  Miles Macklin and
                  Florian Golemo and
                  Vikram Voleti and
                  Linda Petrini and
                  Martin Weiss and
                  Breandan Considine and
                  J{\'{e}}r{\^{o}}me Parent{-}L{\'{e}}vesque and
                  Kevin Xie and
                  Kenny Erleben and
                  Liam Paull and
                  Florian Shkurti and
                  Derek Nowrouzezahrai and
                  Sanja Fidler},
  title        = {gradSim: Differentiable simulation for system identification and visuomotor
                  control},
  booktitle    = {International Conference on Learning Representations},
  year         = {2021},
  url          = {https://openreview.net/forum?id=c\_E8kFWfhp0}
}

@inproceedings{zhang2024physdreamer,
  author       = {Tianyuan Zhang and
                  Hong{-}Xing Yu and
                  Rundi Wu and
                  Brandon Y. Feng and
                  Changxi Zheng and
                  Noah Snavely and
                  Jiajun Wu and
                  William T. Freeman},
  title        = {PhysDreamer: Physics-Based Interaction with 3D Objects via Video Generation},
  booktitle    = {European Conference on Computer Vision},
  volume       = {15060},
  pages        = {388--406},
  publisher    = {Springer},
  year         = {2024},
  url          = {https://doi.org/10.1007/978-3-031-72627-9\_22}
}

@InProceedings{shuai2025pugszeroshotphysicalunderstanding,
  author={Shuai, Yinghao and Yu, Ran and Chen, Yuantao and Jiang, Zijian and Song, Xiaowei and Wang, Nan and Zheng, Jv and Ma, Jianzhu and Yang, Meng and Wang, Zhicheng and Ding, Wenbo and Zhao, Hao},
  booktitle={2025 IEEE International Conference on Robotics and Automation (ICRA)}, 
  title={PUGS: Zero-Shot Physical Understanding with Gaussian Splatting}, 
  year={2025},
  volume={},
  number={},
  pages={4478-4485}
}

@InProceedings{li2024generative,
    author    = {Li, Zhengqi and Tucker, Richard and Snavely, Noah and Holynski, Aleksander},
    title     = {Generative Image Dynamics},
    booktitle = {Proceedings of the IEEE/CVF Conference on Computer Vision and Pattern Recognition},
    month     = {June},
    year      = {2024},
    pages     = {24142-24153}
}

@InProceedings{li2025wonderplay,
    author    = {Li, Zizhang and Yu, Hong-Xing and Liu, Wei and Yang, Yin and Herrmann, Charles and Wetzstein, Gordon and Wu, Jiajun},
    title     = {WonderPlay: Dynamic 3D Scene Generation from a Single Image and Actions},
    booktitle = {Proceedings of the IEEE/CVF International Conference on Computer Vision},
    month     = {October},
    year      = {2025},
    pages     = {9080-9090}
}

@article{cao2025physx3d,
  title={PhysX-3D: Physical-Grounded {3D} Asset Generation},
  author={Ziang Cao and Zhaoxi Chen and Liang Pan and Ziwei Liu},
  journal={arXiv preprint arXiv:2507.12465},
  year={2025}
}

@INPROCEEDINGS{xie2024physgaussian,
  author={Xie, Tianyi and Zong, Zeshun and Qiu, Yuxing and Li, Xuan and Feng, Yutao and Yang, Yin and Jiang, Chenfanfu},
  booktitle={Proceedings of the IEEE/CVF Conference on Computer Vision and Pattern Recognition}, 
  title={PhysGaussian: Physics-Integrated 3D Gaussians for Generative Dynamics}, 
  year={2024},
  volume={},
  number={},
  pages={4389-4398}
  }

@InProceedings{chen2025vid2sim,
    author    = {Chen, Chuhao and Dou, Zhiyang and Wang, Chen and Huang, Yiming and Chen, Anjun and Feng, Qiao and Gu, Jiatao and Liu, Lingjie},
    title     = {Vid2Sim: Generalizable, Video-based Reconstruction of Appearance, Geometry and Physics for Mesh-free Simulation},
    booktitle = {Proceedings of the IEEE/CVF Conference on Computer Vision and Pattern Recognition},
    month     = {June},
    year      = {2025},
    pages     = {26545--26555}
}

@article{modi2024simplicits,
author = {Modi, Vismay and Sharp, Nicholas and Perel, Or and Sueda, Shinjiro and Levin, David I. W.},
title = {Simplicits: Mesh{-}Free, Geometry{-}Agnostic Elastic Simulation},
year = {2024},
issue_date = {July 2024},
publisher = {Association for Computing Machinery},
volume = {43},
number = {4},
issn = {0730-0301},
url = {https://doi.org/10.1145/3658184},
doi = {10.1145/3658184},
journal = {ACM Trans. Graph.},
month = jul,
articleno = {117},
}

@inproceedings{jaegle2021perceiver,
author       = {Andrew Jaegle and
Sebastian Borgeaud and
Jean{-}Baptiste Alayrac and
Carl Doersch and
Catalin Ionescu and
David Ding and
Skanda Koppula and
Daniel Zoran and
Andrew Brock and
Evan Shelhamer and
Olivier J. H{\'{e}}naff and
Matthew M. Botvinick and
Andrew Zisserman and
Oriol Vinyals and
Jo{\~{a}}o Carreira},
title        = {Perceiver {IO:} {A} General Architecture for Structured Inputs {\&}
Outputs},
booktitle    = {International Conference on Learning Representations},
year         = {2022},
url          = {https://openreview.net/forum?id=fILj7WpI-g}
}

@inproceedings{radford2021learning,
  author       = {Alec Radford and
                  Jong Wook Kim and
                  Chris Hallacy and
                  Aditya Ramesh and
                  Gabriel Goh and
                  Sandhini Agarwal and
                  Girish Sastry and
                  Amanda Askell and
                  Pamela Mishkin and
                  Jack Clark and
                  Gretchen Krueger and
                  Ilya Sutskever},
  title        = {Learning Transferable Visual Models From Natural Language Supervision},
  booktitle    = {Proceedings of the 38th International Conference on Machine Learning},
  volume       = {139},
  pages        = {8748--8763},
  publisher    = {{PMLR}},
  year         = {2021}
}

@inproceedings{li2023pac,
  author       = {Xuan Li and
                  Yi{-}Ling Qiao and
                  Peter Yichen Chen and
                  Krishna Murthy Jatavallabhula and
                  Ming Lin and
                  Chenfanfu Jiang and
                  Chuang Gan},
  title        = {PAC-NeRF: Physics Augmented Continuum Neural Radiance Fields for Geometry-Agnostic
                  System Identification},
  booktitle    = {International Conference on Learning Representations},
  year         = {2023},
  url          = {https://openreview.net/forum?id=tVkrbkz42vc}
}

@inproceedings{physctrl2025,
  author    = {Chen Wang and
               Chuhao Chen and
               Yiming Huang and
               Zhiyang Dou and
               Yuan Liu and
               Jiatao Gu and
               Lingjie Liu},
  title     = {PhysCtrl: Generative Physics for 
               Controllable and Physics-Grounded 
               Video Generation},
  booktitle = {Advances in Neural Information Processing Systems},
  year      = {2025}
}

@inproceedings{zhong2024springgaus,
  author       = {Licheng Zhong and
                  Hong{-}Xing Yu and
                  Jiajun Wu and
                  Yunzhu Li},
  title        = {Reconstruction and Simulation of Elastic Objects with Spring-Mass 3D Gaussians},
  booktitle    = {European Conference on Computer Vision},
  volume       = {15060},
  pages        = {407--423},
  publisher    = {Springer},
  year         = {2024},
  url          = {https://doi.org/10.1007/978-3-031-72627-9\_23},
  doi          = {10.1007/978-3-031-72627-9\_23}
}

@InProceedings{jiang2025phystwin,
author    = {Jiang, Hanxiao and Hsu, Hao-Yu and Zhang, Kaifeng and Yu, Hsin-Ni and Wang, Shenlong and Li, Yunzhu},
title     = {PhysTwin: Physics-Informed Reconstruction and Simulation of Deformable Objects from Videos},
booktitle = {Proceedings of the IEEE/CVF International Conference on Computer Vision},
month     = {October},
year      = {2025},
pages     = {7219-7230}
}

@inproceedings{jiang2016mpm,
author = {Jiang, Chenfanfu and Schroeder, Craig and Teran, Joseph and Stomakhin, Alexey and Selle, Andrew},
title = {The material point method for simulating continuum materials},
year = {2016},
publisher = {Association for Computing Machinery},
url = {https://doi.org/10.1145/2897826.2927348},
doi = {10.1145/2897826.2927348},
booktitle = {ACM SIGGRAPH 2016 Courses},
articleno = {24}
}

@inproceedings{lipman2022flow,
author       = {Yaron Lipman and
              Ricky T. Q. Chen and
              Heli Ben{-}Hamu and
              Maximilian Nickel and
              Matthew Le},
title        = {Flow Matching for Generative Modeling},
booktitle    = {International Conference on Learning Representations},
year         = {2023},
url          = {https://openreview.net/forum?id=PqvMRDCJT9t}
}

@article{wang2015dcms,
author = {Wang, Bin and Wu, Longhua and Yin, KangKang and Ascher, Uri and Liu, Libin and Huang, Hui},
title = {Deformation capture and modeling of soft objects},
year = {2015},
issue_date = {August 2015},
publisher = {Association for Computing Machinery},
address = {New York, NY, USA},
volume = {34},
number = {4},
issn = {0730-0301},
url = {https://doi.org/10.1145/2766911},
doi = {10.1145/2766911},
journal = {ACM Trans. Graph.},
month = jul,
articleno = {94},
numpages = {12}
}

@article{wang2004ssim,
author={Zhou Wang and Bovik, A.C. and Sheikh, H.R. and Simoncelli, E.P.},
journal={IEEE Transactions on Image Processing}, 
title={Image quality assessment: from error visibility to structural similarity}, 
year={2004},
volume={13},
number={4},
pages={600-612},
doi={10.1109/TIP.2003.819861}
}

@InProceedings{zhang2018unreasonable,
    author = {Zhang, Richard and Isola, Phillip and Efros, Alexei A. and Shechtman, Eli and Wang, Oliver},
    title = {The Unreasonable Effectiveness of Deep Features as a Perceptual Metric},
    booktitle = {Proceedings of the IEEE/CVF Conference on Computer Vision and Pattern Recognition},
    year={2018},
    volume={},
    number={},
    pages={586-595}
}

@INPROCEEDINGS{xiang2024trellis,
  author={Xiang, Jianfeng and Lv, Zelong and Xu, Sicheng and Deng, Yu and Wang, Ruicheng and Zhang, Bowen and Chen, Dong and Tong, Xin and Yang, Jiaolong},
  booktitle={Proceedings of the IEEE/CVF Conference on Computer Vision and Pattern Recognition}, 
  title={Structured 3D Latents for Scalable and Versatile 3D Generation}, 
  year={2025},
  volume={},
  number={},
  pages={21469-21480}
}
}

\pagebreak
\section*{Appendix}


\section{Dataset Details}
\label{sec:appendix_dataset}

In this section, we provide a comprehensive overview of our new dataset, \dataset. Our work builds upon the 3D assets of the PIXIEVERSE dataset~\cite{le2025pixie} but introduces a fundamentally new annotation paradigm to support our generative and unified modeling goals. Specifically, we re-annotate the entire dataset with \textit{plausible property ranges} and generate consistent parameters for \textit{multiple physics solvers}.

\subsection{Annotation Pipeline for \dataset}
Our semi-automatic annotation pipeline extends the process from PIXIE~\cite{le2025pixie} with two key novelties: (1) annotating a continuous range $[\boldsymbol{y}_{\min}, \boldsymbol{y}_{\max}]$ for each physical property instead of a single point value, and (2) generating consistent ground-truth parameters for LBS and Spring-Mass solvers by fitting them to MPM-driven dynamics.

\begin{figure*}[t]
\centering
\begin{promptbox}[Full VLM Prompt for Range Annotation]
\footnotesize
\textbf{SYSTEM PROMPT: 3D PHYSICS ANNOTATOR} \\
You are UNIPIXIE\_ANNOTATOR, an expert assistant in continuum mechanics and Material Point Method (MPM) simulation.

\vspace{0.3em}
\textbf{\#\#\# TASK OVERVIEW} \\
Given images of a single 3D object, your task is to propose a physically consistent RANGE of material properties rather than a single value. Many objects can appear the same but have different stiffness or density (e.g., dry vs. fresh wood).

\vspace{0.5em}
\textbf{\#\#\# THE CHALLENGE} \\
Objects exist along a soft-to-stiff continuum. Your output must capture this ambiguity with a plausible interval [min, max] for each property. Never output single scalars.

\vspace{0.5em}
\textbf{\#\#\# INPUTS} 
\begin{itemize}
    \setlength\itemsep{0em}
    \item Multi-view RGB images of the target object.
    \item A short semantic category label.
\end{itemize}
Use both geometric and semantic cues.

\vspace{0.5em}
\textbf{\#\#\# REQUIRED OUTPUT (JSON)} \\
You MUST output a JSON object containing:
\begin{itemize}
    \setlength\itemsep{0em}
    \item \texttt{material\_dict}: Per-part ranges for E (Pa), $\rho$ (kg/m$^3$), $\nu$.
    \item \texttt{segmentation\_queries}: CLIP-friendly text queries.
    \item \texttt{reasoning}: Brief explanation of visual cues used.
    \item \texttt{constraints}: Python assert statements.
\end{itemize}

\textbf{1. Semantic Segmentation \& Queries} \\
Decompose the object into FUNCTIONAL parts (`pot', `trunk', `leaves'...). Parts must differ in physical behavior. Provide CLIP-friendly queries such as `ceramic pot' or `woody trunk'.

\vspace{0.3em}
\textbf{2. Material Properties (Plausible Ranges)} \\
For each part, propose [min, max] ranges for:
\begin{itemize}
    \setlength\itemsep{0em}
    \item Young's Modulus E (Pa)
    \item Density $\rho$ (kg/m$^3$)
    \item Poisson's Ratio $\nu$
\end{itemize}
Choose a plausible interval for each property.

\vspace{0.3em}
\textbf{3. Range Design Principles}
\begin{itemize}
    \setlength\itemsep{0em}
    \item Ranges must be plausible and non-empty.
    \item Intervals must create visually distinct soft vs. stiff behavior.
    \item Semantically impossible combinations must be avoided.
\end{itemize}

\vspace{0.3em}
\textbf{4. Pythonic Constraints} \\
Write Python assert statements enforcing global consistency. They must hold for ANY sampled value within each range. \\
Examples:
\begin{itemize}
    \setlength\itemsep{0em}
    \item pot is stiffer \& denser than trunk/leaves
    \item trunk is stiffer than leaves
\end{itemize}

\vspace{0.3em}
\textbf{\#\#\# IN-CONTEXT EXAMPLE (Specific Ficus Tree)} \\
\textit{Input: A bonsai with a thick, rough bark trunk and a heavy unglazed ceramic pot.} \\
Assistant Output:
\begin{lstlisting}[language=json]
{
  "material_dict": {
    "pot": {
      "density": [1800, 2400], "E": [1e9, 5e9], "nu": [0.15, 0.2]
    },
    "trunk": {
      "density": [600, 800], "E": [5e6, 5e7], "nu": [0.3, 0.35]
    },
    "leaves": {
      "density": [200, 400], "E": [1e4, 5e4], "nu": [0.4, 0.45]
    }
  },
  "reasoning": "Thick lignified trunk suggests high stiffness for wood. Unglazed clay pot implies high density.",
  "constraints": "assert mat['pot']['E'][0] > mat['trunk']['E'][1]"
}
\end{lstlisting}
\end{promptbox}
\caption{\textbf{Full VLM Prompt for Physical Property Range Annotation.} We provide the VLM with detailed system instructions, task definitions, and in-context examples (JSON format) to guide it in generating plausible physical property ranges and constraints.}
\label{fig:appendix_vlm_prompt_text}
\end{figure*}

\begin{figure*}[t]
    \centering
    \includegraphics[width=0.88\textwidth]{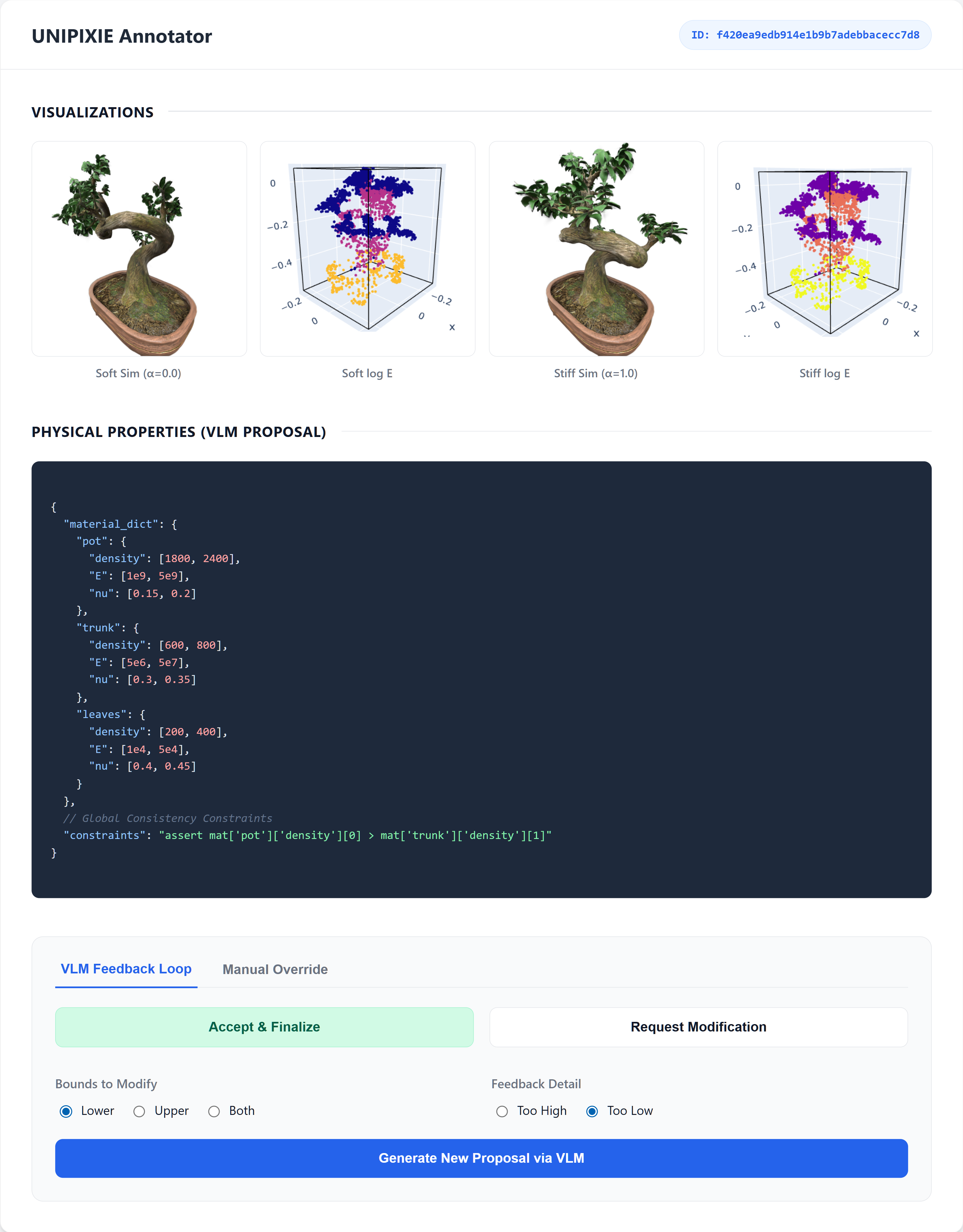}
    \caption{\textbf{Interactive Interface for Manual Verification and Refinement.} Our web-based platform ensures high-quality annotations for \dataset. It presents side-by-side visualizations (simulation videos and log E maps) of the soft ($\boldsymbol{y}_{\min}$) and stiff ($\boldsymbol{y}_{\max}$) endpoints alongside the VLM's proposal. Experts can either (a) Accept the proposal, (b) Request Modification via structured feedback to trigger a VLM revision loop, or (c) Manually Override specific range values directly. This flexible, human-in-the-loop workflow guarantees both efficiency and physical plausibility.}
\label{fig:appendix_annotation_interface}
\end{figure*}

\subsubsection{MPM Range Annotation}
\label{sec:appendix_mpm_range}
We employ a two-stage pipeline to annotate plausible physical ranges for the Material Point Method (MPM) solver, combining VLM priors with rigorous human verification.

\paragraph{VLM-based Range Proposal.}
Extending the pipeline from PIXIE~\cite{le2025pixie}, we prompt a Vision-Language Model (VLM) to propose a plausible \textit{range} $[\boldsymbol{y}_{\min}, \boldsymbol{y}_{\max}]$ for each physical property (Young's modulus, density, Poisson's ratio) instead of a single point estimate. As detailed in Figure~\ref{fig:appendix_vlm_prompt_text}, the prompt instructs the VLM to perform visual-to-physical reasoning (e.g., inferring stiffness from texture) and specify inter-part constraints (e.g., trunk $>$ leaves), leveraging its world knowledge to initialize physically grounded ranges.

\paragraph{Interactive Verification and Refinement.}
To ensure data quality, we developed a annotation interface (Figure~\ref{fig:appendix_annotation_interface}) for human-in-the-loop refinement. Annotators evaluate the VLM proposals based on simultaneous visualizations of the soft ($\boldsymbol{y}_{\min}$) and stiff ($\boldsymbol{y}_{\max}$) endpoint simulations. The workflow consists of:
\begin{enumerate}
    \item \textbf{Evaluation:} Annotators check for Physical Plausibility (no artifacts/explosions) and Visual Diversity (distinct behaviors at endpoints).
    \item \textbf{Feedback Loop:} If the range is unsatisfactory, annotators specify the necessary adjustment (e.g., ``Lower bound is too high").
    \item \textbf{Refinement:} Annotators can either manually adjust the range values directly or request an auto-refinement, where the system queries the VLM with the feedback to generate a revised proposal. This cycle repeats until the range is accepted.
\end{enumerate}

\subsubsection{Cross-Solver Parameter Generation}
\label{sec:appendix_cross_solver}
To train our unified architecture, we generate ground-truth parameters for Linear Blend Skinning (LBS) and Spring-Mass solvers that are dynamically consistent with our primary MPM annotations. We adopt an MPM-centric approach, fitting parameters to match the dynamics of MPM simulations at $\alpha \in \{0, 1\}$.

\paragraph{Linear Blend Skinning (LBS).}
We use the test-time optimization framework from Vid2Sim~\cite{chen2025vid2sim} to fit LBS parameters to 30-frame MPM videos rendered at the soft and stiff endpoints.
\begin{itemize}
    \item \textbf{Deformation Model:} We observe that the underlying skinning weights and control handles remain nearly identical across stiffness levels. Thus, we simplify the ground truth by using a single LBS model fitted to the softest ($\boldsymbol{y}_{\min}$) video as the object-specific deformation basis.
    \item \textbf{Material Properties:} The continuous range is controlled solely by varying the material parameters (Young's Modulus, Poisson's Ratio), which are fitted to match the endpoint dynamics and verified via our interactive interface.
\end{itemize}

\paragraph{Spring-Mass System.}
For the Spring-Mass system, we aim to learn intrinsic stiffness properties decoupled from extrinsic factors.
\begin{itemize}
    \item \textbf{Base Model:} We fit a Spring-Gaus~\cite{zhong2024springgaus} model to the softest MPM video (30 frames) to determine anchor points and spring topology. Extrinsic parameters (damping, initial velocity) are fixed globally.
    \item \textbf{Softness Vector:} To capture the soft-to-stiff distribution, we focus on the \textit{softness vector} $\boldsymbol{\eta}$. Instead of re-fitting the entire model, annotators use our platform to define a plausible range for $\boldsymbol{\eta}$ that modulates the global stiffness to match the visual dynamics of the MPM simulations.
\end{itemize}

\section{Model Architecture and Training Details}
\label{sec:appendix_model}

In this section, we provide a comprehensive specification of the \method architecture, training objectives, hyperparameters, and inference procedures to ensure full reproducibility. 

\subsection{Detailed Model Architectures}
Our framework comprises a shared Grid Encoder and a suite of specialized decoders tailored for different physics solvers. A summary of the network configurations is provided in Table~\ref{tab:network_configs}.

\begin{table*}[h]
  \centering
  \small
  \caption{\textbf{Network Configurations.} Summary of architecture hyperparameters for the unified encoder and specialized decoders.}
  \label{tab:network_configs}
  \begin{tabular}{l c c c l l}
    \toprule
    \textbf{Module} & \textbf{\# Layers} & \textbf{Dim ($C$)} & \textbf{\# Heads} & \textbf{Block Architecture} & \textbf{Special Mechanisms} \\
    \midrule
    \textbf{Grid Encoder} & 6 & 512 & 8 & Cross-Attn + Self-Attn & - \\
    \midrule
    \textbf{Main FMT (MPM)} & 6 & 512 & 8 & FMT Block (Cross-Attn + MLP) & AdaLN-Zero, QK Norm, SwiGLU \\
    \textbf{LBS (HyperNet)} & 4 & 512 & - & MLP & - \\
    \textbf{LBS (Material)} & 6 & 512 & 8 & FMT Block (Shared w/ Main) & AdaLN-Zero, QK Norm, SwiGLU \\
    \textbf{Spring-Mass} & 4 & 512 & 8 & FMT Block (Vector) & AdaLN-Zero, QK Norm, SwiGLU \\
    \bottomrule
  \end{tabular}
\end{table*}

\subsubsection{Grid Encoder}
The Grid Encoder $\mathcal{E}$ distills the high-dimensional voxelized CLIP features $\mathcal{G}_{\text{feat}} \in \mathbb{R}^{64^3 \times 768}$ into a compact set of $L=64$ latent tokens $\boldsymbol{z}_{\text{latent}}$. We employ a tokenizer-based architecture inspired by Perceiver-IO~\cite{jaegle2021perceiver}.
\begin{itemize}
    \item \textbf{Convolutional Stem:} A 3D convolutional stem (2 blocks of Conv3d $4{\times}4{\times}4$, stride 2) downsamples the input from $64^3$ to $16^3$ while mapping channels to the latent dimension $C=512$.
    \item \textbf{Latent Tokenizer:} The learnable latent tokens query the flattened grid features via 6 Transformer blocks. We use Fourier Positional Encodings (16 frequencies) to embed 3D coordinates.
\end{itemize}

\subsubsection{Flow Matching Transformer (FMT) Decoder}
The core generative backbone is the Flow Matching Transformer (FMT). It conditions on the latent tokens $\boldsymbol{z}_{\text{latent}}$, timestep $t$, and control parameter $\alpha$ to predict the velocity field.

\paragraph{Unified Conditioning.}
We employ a fused conditioning scheme. The scalar timestep $t$ and control parameter $\alpha$ are embedded via Fourier features and sinusoidal embeddings, respectively. These are concatenated with the global average of $\boldsymbol{z}_{\text{latent}}$ and fused via an MLP into a single vector $\boldsymbol{c}$.

\paragraph{Transformer Block Design.}
The decoder consists of $N=6$ blocks designed for stable generative modeling. As detailed in our implementation, each block contains two primary sub-layers: a Cross-Attention layer and a Feed-Forward Network (MLP).
\begin{itemize}
    \item \textbf{Cross-Attention:} The input features serve as queries, while the global latent tokens $\boldsymbol{z}_{\text{latent}}$ serve as keys and values. This allows the decoder to attend to the physics-aware global context at every layer.
    \item \textbf{Adaptive Layer Normalization (AdaLN):} The fused condition $\boldsymbol{c}$ modulates the normalization layers. We regress scale ($\gamma$), shift ($\beta$), and gate ($\alpha$) parameters for each sub-layer from $\boldsymbol{c}$.
    \item \textbf{AdaLN-Zero Initialization:} We initialize the final gating weights of each block to zero. This effectively initializes the block as an identity function, which has been shown to improve training stability and convergence.
    \item \textbf{QK Normalization:} To prevent attention instability, we apply RMSNorm to the Queries and Keys within the attention layers.
    \item \textbf{SwiGLU Activation:} The feed-forward networks utilize the SwiGLU activation function for improved expressivity.
\end{itemize}

\subsubsection{Specialized Decoder Heads}
All decoders branch from the same shared latent representation $\boldsymbol{z}_{\text{latent}}$, but utilize architectures specifically tailored to the input requirements of their respective physics engines.

\begin{itemize}
    \item \textbf{MPM Decoder:} Integrated directly into the main FMT backbone. It operates on the voxel grid and outputs $3$ continuous channels (log-Young's modulus, log-density, Poisson's ratio) and $8$ categorical channels (material logits) per voxel.
    
    \item \textbf{LBS Decoder (Dual-Branch):} For Linear Blend Skinning (LBS), we adopt the framework of Vid2Sim~\cite{chen2025vid2sim}, which decouples the deformation model (skinning weights) from material properties. Accordingly, our LBS decoder consists of two branches:
    \begin{enumerate}
        \item \textbf{Deformation Model (HyperNetwork):} Following Vid2Sim's strategy, we employ a HyperNetwork implemented as a 4-layer MLP. It takes the global shape embedding (averaged $\boldsymbol{z}_{\text{latent}}$) and directly regresses the flattened parameters $\theta_{\text{LBS}}$ (weights and biases, total dim 650) of the LBS deformation network.
        \item \textbf{Material Properties (FMT Head):} To predict the spatially-varying material fields (Young's Modulus and Poisson's Ratio) required by the simulator, we use a grid-based Flow Matching head. This head shares the identical FMT backbone structure as the MPM decoder described above, ensuring consistent generative modeling across solvers.
    \end{enumerate}
    
    \item \textbf{Spring-Mass Decoder:} For the particle-based Spring-Mass system, we build upon the Spring-Gaus~\cite{zhong2024springgaus} representation. Spring-Gaus introduces a global \textit{softness vector} to efficiently modulate the stiffness of the entire system. Adopting this efficient parameterization, our decoder utilizes a separate Vector Flow Matching Decoder. It shares the same transformer block design as our main FMT but operates on global vectors. It predicts a 2049-dimensional vector $\boldsymbol{m}_{\text{spring}}$, concatenating the baseline stiffness parameters with the global softness vector $\boldsymbol{\eta}$.
\end{itemize}

\subsection{Training Objectives and Hyperparameters}
We train the entire framework end-to-end. The total loss is a weighted sum of task-specific losses and regularization terms:
\begin{equation}
    \mathcal{L}_{\text{total}} = \mathcal{L}_{\text{MPM}} + \lambda_{\text{LBS}}\mathcal{L}_{\text{LBS}} + \lambda_{\text{SM}}\mathcal{L}_{\text{SM}} + \lambda_{\text{KL}}\mathcal{L}_{\text{KL}}
\end{equation}

\paragraph{Loss Functions.}
\begin{itemize}
    \item \textbf{Conditional Flow Matching (CFM):} For MPM, LBS Material head, and Spring-Mass, we use the CFM objective. We sample a target $\boldsymbol{x}_1$ (interpolated via LERP), noise $\boldsymbol{x}_0 \sim \mathcal{N}(0, I)$, and timestep $t$ from a logit-Normal distribution. The model predicts the velocity field $\boldsymbol{v}_\theta$, optimized via MSE against $\boldsymbol{x}_1 - \boldsymbol{x}_0$.
    \item \textbf{LBS Regression:} For the HyperNetwork branch, we use MSE loss between the predicted LBS parameters and the interpolated ground-truth parameters.
    \item \textbf{KL Regularization:} We apply a KL Consistency loss ($\lambda=1e^{-4}$) that enforces consistency between the latent embeddings of the original input and a perturbed input (noise level 0.01).
\end{itemize}

\paragraph{Hyperparameters.}
The model is trained on 4 NVIDIA A6000 GPUs. Detailed settings are provided in Table~\ref{tab:appendix_hyperparams}.

\begin{table}[h]
  \centering
  \small
  \caption{\textbf{Training Hyperparameters.}}
  \label{tab:appendix_hyperparams}
  \begin{tabular}{lc}
    \toprule
    \textbf{Parameter} & \textbf{Value} \\
    \midrule
    Batch Size & 1 per GPU (Effective: 4) \\
    Optimizer & AdamW \\
    Learning Rate & $5 \times 10^{-5}$ \\
    Weight Decay & $0.01$ \\
    LR Scheduler & Cosine Annealing (Warmup: 3000 steps) \\
    Training Epochs & 100 \\
    Gradient Clipping & 1.0 \\
    Mixed Precision & FP16 (Enabled) \\
    \midrule
    $\lambda_{\text{LBS}}$ & 0.8 \\
    $\lambda_{\text{SM (Stiffness)}}$ & 0.5 \\
    $\lambda_{\text{SM (Softness Vector)}}$ & 2.5 \\
    \bottomrule
  \end{tabular}
\end{table}

\subsection{Inference Procedure}
At inference time, we generate physical parameters by solving the probability flow ODE defined by our learned vector field $\boldsymbol{v}_\theta$. We employ Heun's Method, a second-order Runge-Kutta numerical solver, to integrate from time $t=0$ (noise) to $t=1$ (data).

Given a noise sample $\boldsymbol{x}_0 \sim \mathcal{N}(0, I)$ and a step size $\Delta t = 1/N$ (where $N=15$ is the number of inference steps), each update step from $t_i$ to $t_{i+1}$ consists of a predictor and a corrector:

\begin{align}
    \text{\textbf{Predictor:}} \quad \tilde{\boldsymbol{x}}_{i+1} &= \boldsymbol{x}_i + \boldsymbol{v}_\theta(\boldsymbol{x}_i, t_i) \Delta t \\
    \text{\textbf{Corrector:}} \quad \boldsymbol{x}_{i+1} &= \boldsymbol{x}_i + \frac{\Delta t}{2} \left[ \boldsymbol{v}_\theta(\boldsymbol{x}_i, t_i) + \boldsymbol{v}_\theta(\tilde{\boldsymbol{x}}_{i+1}, t_{i+1}) \right]
\end{align}

This 2nd-order approximation provides a superior trade-off between generation quality and computational cost compared to the standard Euler method, requiring only 30 function evaluations (NFE) for high-fidelity results.

\section{Baseline Implementation Details}
\label{sec:appendix_baselines}

To ensure a fair and comprehensive evaluation of \method, we carefully adapted and re-trained all baseline methods on our \dataset. This section details the specific implementation and training protocols for each baseline.

\subsection{Deterministic Baselines}
\label{sec:appendix_deterministic_baselines}

\paragraph{PIXIE~\cite{le2025pixie}.}
As PIXIE is the direct predecessor to our work, we treat it as our primary deterministic baseline.
\begin{itemize}
    \item \textbf{Re-training:} We conducted a full re-training of the official PIXIE model on our \dataset. To adapt it to our dataset's format, we used the midpoint of our annotated property ranges (i.e., parameters generated with $\alpha=0.5$) as the single ground-truth target for its supervised loss.
    \item \textbf{Hyperparameters:} We used the official open-source implementation and followed the hyperparameter settings reported in the original paper, including the U-Net architecture, learning rate, and optimizer settings, to ensure a faithful comparison.
\end{itemize}

\paragraph{NeRF2Physics~\cite{zhai2024physical} and PUGS~\cite{shuai2025pugszeroshotphysicalunderstanding}.}
These methods leverage Vision-Language Models (VLMs) for zero-shot physics prediction. While PUGS originally supports continuous material properties (e.g., density, Young's modulus), it does not natively predict discrete material IDs required for physics simulation. To obtain material IDs, we extended PUGS by creating a specialized prompt that instructs GPT-4V to classify materials into 7 discrete categories based on their physical behavior: jelly (0) for deformable materials like rubber and elastic bands, metal (1) for metallic materials, sand (2) for granular materials, visplas (3) for viscoelastic plastics like clay and putty, fluid (4) for liquids, snow (5) for snow and ice-like materials, and stationary (6) for rigid, non-deformable materials. The prompt asks the VLM to analyze the input image and output a JSON response containing material names paired with their corresponding material IDs (0-6), enabling downstream physics simulation with appropriate material models.


\subsection{Generative Ablation Baseline}
\label{sec:appendix_ablation_baseline}

\paragraph{3D U-Net (Ablation).}
To validate the superiority of our Transformer-based architecture, we compare against a strong generative baseline based on a 3D U-Net.
\begin{itemize}
    \item \textbf{Architecture:} This baseline replaces our Flow Matching Transformer (FMT) with a conditional 3D U-Net. The network takes the noisy material grid and projected visual features as input. It utilizes a standard encoder-decoder structure with residual blocks, attention layers at lower resolutions.
    \item \textbf{Conditioning:} Similar to our full model, the U-Net is conditioned on the control parameter $\alpha$. We map $\alpha$ to a sinusoidal embedding, which modulates the residual blocks via Adaptive Group Normalization (AdaGN).
    \item \textbf{Training \& Inference:} Unlike the flow matching objective of our main model, this baseline is trained as a standard Denoising Diffusion Probabilistic Model (DDPM). At inference time, we employ the Denoising Diffusion Implicit Models (DDIM) sampler with 50 steps ($\eta=0.0$) to generate the material fields, ensuring a fair comparison of generative capabilities.
\end{itemize}

\subsection{Specialized Solver Baselines}
\label{sec:appendix_specialized_baselines}
For our multi-solver evaluation, we compare against state-of-the-art methods specialized for each physics backend.

\paragraph{Vid2Sim~\cite{chen2025vid2sim}.}
We use the official implementation of Vid2Sim for the LBS solver comparison.
\begin{itemize}
    \item \textbf{Vid2Sim (full):} This refers to the standard model running its test-time optimization procedure for the number of iterations specified in the original paper to achieve the best performance.
    \item \textbf{Vid2Sim (fast):} To provide a baseline with a more comparable runtime to feed-forward methods, we created an accelerated variant. This version runs the same optimization procedure but for only one-third of the iterations of the ``full" version.
\end{itemize}

\paragraph{Spring-Gaus~\cite{zhong2024springgaus}.}
We compare against the official implementation for the Spring-Mass system.
\begin{itemize}
    \item \textbf{Spring-Gaus:} This is the original model running its standard test-time optimization.
    \item \textbf{Spring-Gaus (tuned):} The original Spring-Gaus model has several hyperparameters that can be tuned for a specific data distribution. To create the strongest possible baseline, we performed a hyperparameter search on a small validation split of our \dataset. The ``tuned" version uses these optimized hyperparameters (e.g., for spring initialization and damping), resulting in better performance on our specific set of elastic objects compared to the default configuration.
\end{itemize}

\end{document}